\documentclass[twocolumn,aps,prb,10pt,superscriptaddress,longbibliography,floatfix,citeautoscript]{revtex4-2}
\usepackage{amsmath}
\usepackage{amssymb}
\usepackage{bm}
\usepackage{dcolumn}
\usepackage{glossaries}
\usepackage{graphicx}
\usepackage{multirow}
\usepackage{fancyvrb}
\usepackage{siunitx}
\usepackage[usenames,dvipsnames,svgnames]{xcolor}
\usepackage{hyperref}
\hypersetup{
pdfnewwindow=true,      
colorlinks=true,        
linkcolor=Blue,         
citecolor=Blue,         
filecolor=Blue,         
urlcolor=Blue           
}

\renewcommand{\thesection}{\arabic{section}} 
\renewcommand{\thesubsection}{\thesection.\arabic{subsection}} 
\renewcommand{\thesubsubsection}{\thesubsection.\arabic{subsubsection}} 

\usepackage{titlesec}
\titleformat{\section}{\bfseries\large}{\thesection.}{1em}{} 
\titleformat{\subsection}{\bfseries\normalsize}{\thesubsection.}{1em}{} 
\titleformat{\subsubsection}{\bfseries\small}{\thesubsubsection.}{1em}{} 

\usepackage{chemformula} 
\usepackage{mhchem} 
\usepackage[T1]{fontenc} 
\usepackage{xcolor} 

\usepackage[normalem]{ulem}
\usepackage[draft]{changes} 

\newacronym{dft}{DFT}{density functional theory}
\newacronym{eam}{EAM}{embedded atom method}
\newacronym{hnemd}{HNEMD}{homogeneous non-equilibrium molecular dynamics}
\newacronym{mae}{MAE}{mean absolute error}
\newacronym{md}{MD}{molecular dynamics}
\newacronym{mlp}{MLP}{machine-learned potential}
\newacronym{nep}{NEP}{neuroevolution potential}
\newacronym{rmse}{RMSE}{root mean square error}
\newacronym{r2}{$R^2$}{coefficient of determination}
\newacronym{snes}{SNES}{separable natural evolution strategy}
\newacronym{vac}{VAC}{velocity auto-correlation}
\newacronym{vdos}{VDOS}{vibrational density of states}
\newacronym{sdc}{SDC}{self-diffusion coefficient}
\newacronym{gpumd}{GPUMD}{graphics processing units molecular dynamics}
\newacronym{vasp}{VASP}{Vienna ab initio simulation package}
\newacronym{pca}{PCA}{Principal Component Analysis}
\newacronym{umap}{UMAP}{Uniform Manifold Approximation and Projection}
\newacronym{soap}{SOAP}{Smooth Overlap of Atomic Positions}
\newacronym{pbe}{PBE}{Perdew-Burke-Ernzerhof}
\newacronym{eos}{EOS}{equation of state}
\newacronym{hpc}{HPC}{high-performance computing}
\begin{document}

\title{NepTrain and NepTrainKit: Automated Active Learning and Visualization Toolkit for Neuroevolution Potentials}

\author{Chengbing Chen}
\thanks{These authors contributed equally to this work.}
\affiliation{College of physics and electronic information engineering, Guilin University of Technology, Guilin, 541008, P. R. China}

\author{Yutong Li}
\thanks{These authors contributed equally to this work.}
\affiliation{School of Interdisciplinary Science, Beijing Institute of Technology, Beijing 100081, P. R. China}

\author{Rui Zhao}
\email{zrtata@hnu.edu.cn}
\affiliation{School of Mechanical and Electrical Engineering, Xinyu University, Xinyu, 338004, P. R. China}

\author{Zhoulin Liu}
\affiliation{School of Science, Harbin Institute of Technology (Shenzhen), Shenzhen 518055, China}

\author{Zheyong Fan}
\affiliation{College of Physical Science and Technology, Bohai University, Jinzhou 121013, P. R. China}

\author{Gang Tang}
\email{gtang@bit.edu.cn}
\affiliation{School of Interdisciplinary Science, Beijing Institute of Technology, Beijing 100081, P. R. China}

\author{Zhiyong Wang}
\email{zhiyongwang@glut.edu.cn}
\affiliation{College of physics and electronic information engineering, Guilin University of Technology, Guilin, 541008, P. R. China}

\date{\today}

\begin{abstract}
As a machine-learned potential, the neuroevolution potential (NEP) method features exceptional computational efficiency and has been successfully applied in materials science. Constructing high-quality training datasets is crucial for developing accurate NEP models. However, the preparation and screening of NEP training datasets remain a bottleneck for broader applications due to their time-consuming, labor-intensive, and resource-intensive nature. In this work, we have developed NepTrain and NepTrainKit, which are dedicated to initializing and managing training datasets to generate high-quality training sets while automating NEP model training. NepTrain is an open-source Python package that features a bond length filtering method to effectively identify and remove non-physical structures from molecular dynamics trajectories, thereby ensuring high-quality training datasets. NepTrainKit is a graphical user interface (GUI) software designed specifically for NEP training datasets, providing functionalities for data editing, visualization, and interactive exploration. It integrates key features such as outlier identification, farthest-point sampling, non-physical structure detection, and configuration type selection. The combination of these tools enables users to process datasets more efficiently and conveniently. Using $\rm CsPbI_3$ as a case study, we demonstrate the complete workflow for training NEP models with NepTrain and further validate the models through materials property predictions. We believe this toolkit will greatly benefit researchers working with machine learning interatomic potentials.
\end{abstract}

\maketitle

\section{Introduction}

The rapid development of \glspl{mlp}, or machine-learned force fields \cite{Unke2021Machine}, has provided a revolutionary solution for \gls{md} simulations, offering both the accuracy of \gls{dft} calculations and the efficiency of empirical potential. This advancement is driving the shift in computational materials science from empirical models to a data-driven paradigm. Traditional \gls{md} simulations are fundamentally dependent on the accuracy of interatomic potentials, which are typically derived from empirical formulations. However, these empirical potentials often exhibit limitations in describing complex material properties or non-equilibrium states \cite{Behler2021Machine}, hindering a comprehensive capture of the associated physical phenomena. In recent years, the development of \glspl{mlp} has offered a promising approach to address these challenges. Compared to conventional empirical potentials, \gls{mlp} not only demonstrate higher computational accuracy but also exhibit adaptability across diverse material types, demonstrating significant potential for broad applications \cite{Friederich2021Machine-Learned, Behler2016Perspective}.

Many \gls{mlp} methods have been developed and widely used, including, e.g., the Behler-Parrinello neural network potential \cite{behler2007generalized}, the Gaussian approximation potential (GAP) \cite{bartok2010gaussian}, the spectral neighbor analysis potential \cite{Thompson2015jcp}, the moment tensor potential (MTP) \cite{shapeev2016moment}, the deep potential \cite{Wang2018DeePMD-kit,Zhang2018Deep}, the atomic cluster expansion potential \cite{drautz2019atomic}, and the \gls{nep} \cite{Fan2021Neuroevolution}.
Among these \glspl{mlp}, the \gls{nep} approach is known for its exceptional computational efficiency, enabled by its optimized formulation and efficient implementation in the \gls{gpumd} package \cite{Fan2022GPUMD, fan2017efficient}.
Due to its high computational efficiency, the \gls{nep} approach has been extensively used in \gls{md} simulations, addressing complex challenges that traditional force fields typically can not tackle. Notably, the \gls{nep} method has been successfully applied in a wide range of material systems and research areas  \cite{dong2024molecular, ying2025advances}, such as predicting the structural properties of liquid and amorphous materials\cite{Wang2024Density,Xu2024NEPMBpol}, unraveling chemical order in complex alloy systems\cite{Chen2024Intricate}, exploring phase transitions\cite{Fransson2023Phase,Wiktor2023Quantifying}, capturing surface reconstructions and material growth\cite{Qian2024Reconstruction}, simulating primary radiation damage\cite{Liu2023Largescale}, investigating fracture in two-dimensional materials\cite{Ying2023Atomistic,Yu2024Fracture}, nanoscale tribology\cite{ying2025advances}, as well as elucidating the mechanical behavior of compositionally complex alloys under various loading conditions\cite{Song2024General}.

Currently, a variety of tools have been developed for \glspl{mlp}, offering functionalities such as training, evaluation, inference, fine-tuning, and deployment. For example, metatrain is a command-line interface (CLI) for training and evaluating atomistic models with diverse architectures, including the GAP \cite{Metatensor2025Metatrain}. apax is a flexible and efficient open-source software package for both training and inference of \glspl{mlp}, such as the Gaussian Moment Neural Network model \cite{Schafer2025Apax}. equitrain \cite{Philipp2025Equitrain} provides a unified framework for training and fine-tuning \glspl{mlp}. Similarly, mlip is a Python library for training and deploying \glspl{mlp} \cite{Brunken2025MachineLearning}.

In addition to these general-purpose toolkits, several automated frameworks have recently emerged to facilitate the generation of \glspl{mlp}. For instance, autoplex provides a fully automated solution for creating high-quality \glspl{mlp}, with support for the GAP \cite{Liu2024Automated}. DP-GEN is dedicated to the automated generation and training of deep potential models \cite{Zhang2020DPGEN}. AtomProNet\cite{Galib2025AtomProNet} is an open-source Python package that automates the acquisition of atomic structures, the preparation and submission of ab initio calculations, and the efficient collection of batch-processed data for streamlined neural network (NN) training, supporting models such as MACE\cite{Batatia2022Mace}.

 However, existing tools for training \glspl{mlp} rarely support architectures such as the \gls{nep}, and automated frameworks for fully training \gls{nep} models remain scarce. Notably, a key aspect in the development of \glspl{mlp} is the generation of high-quality datasets for model training. Although recent toolkits such as calorine \cite{lindgren2024calorine} provide approaches for generating initial training sets, including strained, deformed, and rattled structures that can be used for the construction and training of \gls{nep} models, the preparation of \gls{nep} training datasets still heavily depends on user expertise. At present, few tools are available that can both systematically construct high-quality training structures and enable automated \gls{nep} model training. Moreover, graphical user interface (GUI) solutions for efficient screening and visualization of high-quality training datasets are also lacking. Addressing these limitations is the focus of our work, aiming to bridge this gap and further promote the broader application of \gls{nep} models. 
 
 Therefore, in this work, we have developed NepTrain and NepTrainKit. NepTrain integrates the complete workflow, including dataset creation, training, and automated active learning, with built-in logic functions to constrain force ranges and filter out non-physical structures based on bond lengths. Furthermore, it allows seamless integration with Python and Bash scripts, enabling user-defined training workflows while automating processes through the use of \gls{nep}, \gls{gpumd}, \gls{vasp} \cite{Kresse1994AbInitio, Kresse1996Efficient}, and other tools. In parallel, NepTrainKit provides a user-friendly and powerful interactive GUI that enables researchers to visually inspect and edit training datasets. Its design aims to streamline the dataset construction process, enhance efficiency, and ensure the accuracy and reliability of potentials. The intuitive operation of NepTrainKit has significantly advanced the development of \gls{nep} and accelerates the study of material properties.

This paper is organized as follows. In Section \ref{sec:nep_gpumd}, we introduce the \gls{nep} method and its implementation within the \gls{gpumd} package. In Section \ref{sec:neptrain}, we provide a detailed introduction to the workflow and core functionalities of each module in NepTrain. Section \ref{sec:neptrainkit} delves into NepTrainKit, including its user interface design, interactive features, and how it assists users in efficiently editing and managing training datasets. In Section \ref{sec:example}, we will present concrete examples, demonstrating the process of building training datasets from scratch and performing a comparative analysis with existing works. These examples aim to help users gain a deeper understanding of the practical applications, effectiveness, and advantages of NepTrain and NepTrainKit.

\section{The NEP method and the GPUMD package}
\label{sec:nep_gpumd}

The \gls{nep} method \cite{Fan2021Neuroevolution, Fan2022GPUMD, song2024general} is a neural network potential which is trained using the separable natural evolution strategy \cite{schaul2011high}.
Like many other neural network potentials \cite{Behler2021Machine}, \gls{nep} is a many-body potential. 
However, it is still a localized potential with well defined site energy $U_i$ for each atom $i$ in an extended system.
The site energy is a function of an abstract descriptor vector $\mathbf{q}^i$ with a number of components $q^i_{\nu}$ ($\nu=1,2,\cdots,N_{\rm des}$). 
Each descriptor component characterizes the structural and chemical environments of the atom $i$ partially.
Similar to the Behler-Parrinello \cite{behler2007generalized} approach, the descriptor components are divided into radial and angular descriptors.
Details on these descriptors are presented in previous works \cite{Fan2021Neuroevolution, Fan2022GPUMD}.
Currently, only a singe hidden layer is used in the neural network model for NEP, and the site energy can be explicitly written as
\begin{equation}
\label{equation:Ui}
U_i = \sum_{\mu=1} ^{N_\mathrm{neu}} w ^{(1)} _{\mu} \tanh\left(\sum_{\nu=1} ^{N_\mathrm{des}} w ^{(0)}_{\mu\nu} q^i_{\nu} - b^{(0)}_{\mu}\right) - b^{(1)}.
\end{equation}
Here, $\tanh(x)$ is the activation function, $w^{(0)}$ are the weight parameters connecting the input layer (with dimension $N_{\rm des}$) and the hidden layer (with dimension $N_{\rm neu}$), $w^{(1)}$ represents the weight parameters connecting the hidden layer and the output layer (the site energy), $b^{(0)}$ represent the bias parameters in the hidden layer, and $b^{(1)}$ is the bias parameter in the output layer. The trainable parameters are optimized by minimizing a loss function, which is constructed as a sum of weighted root-mean-square errors of energy, force, and virial, along with regularization terms.

Apart from the parameters in the neural network, there are also other trainable parameters in the descriptors \cite{Fan2022GPUMD}.
The \gls{gpumd} 4.0 package released recently contain the NEP89 foundation model \cite{liang2025nep89} covering virtually the entire periodic table. 
This pre-trained model provides a way of calculating the descriptors for numerous materials. 

The \gls{nep} method is implemented into the \gls{gpumd} package \cite{fan2017efficient}, which can be used to train \gls{nep} models and perform \gls{md} simulation with the trained models.
Both the training and inference are accelerated by using GPUs, which attain high computational efficiency.
Upon compilation of the \gls{gpumd} source code, two executables are generated: \texttt{gpumd} and \texttt{nep}.
The \texttt{nep} executable is used for training of \gls{nep} models, and the \texttt{gpumd} executable is used for performing \gls{md} simulations.
Both executables need some input files.
The \texttt{nep} executable requires at least a \texttt{nep.in} input file containing the various hyperparameters for training, and a \texttt{train.xyz} file containing the training data.
The \texttt{gpumd} executable requires a \texttt{run.in} file controlling the \gls{md} simulation process and a \texttt{model.xyz} file defining the simulation system.
Both the\texttt{train.xyz} and \texttt{model.xyz} files follow the extended XYZ file format. 
Apart from the GPU implementation in the \gls{gpumd} package, \gls{nep} also has a CPU implementation denoted \texttt{NEP\_CPU} \cite{nepcpu}, which works as a \gls{nep} calculator in CPU platform. 
This CPU implementation of \gls{nep} is used in NepTrain and NepTrainKit, as well as other packages such as calorine \cite{lindgren2024calorine}.

\begin{figure*}[!]
    \centering
    \includegraphics[width=0.8\textwidth]{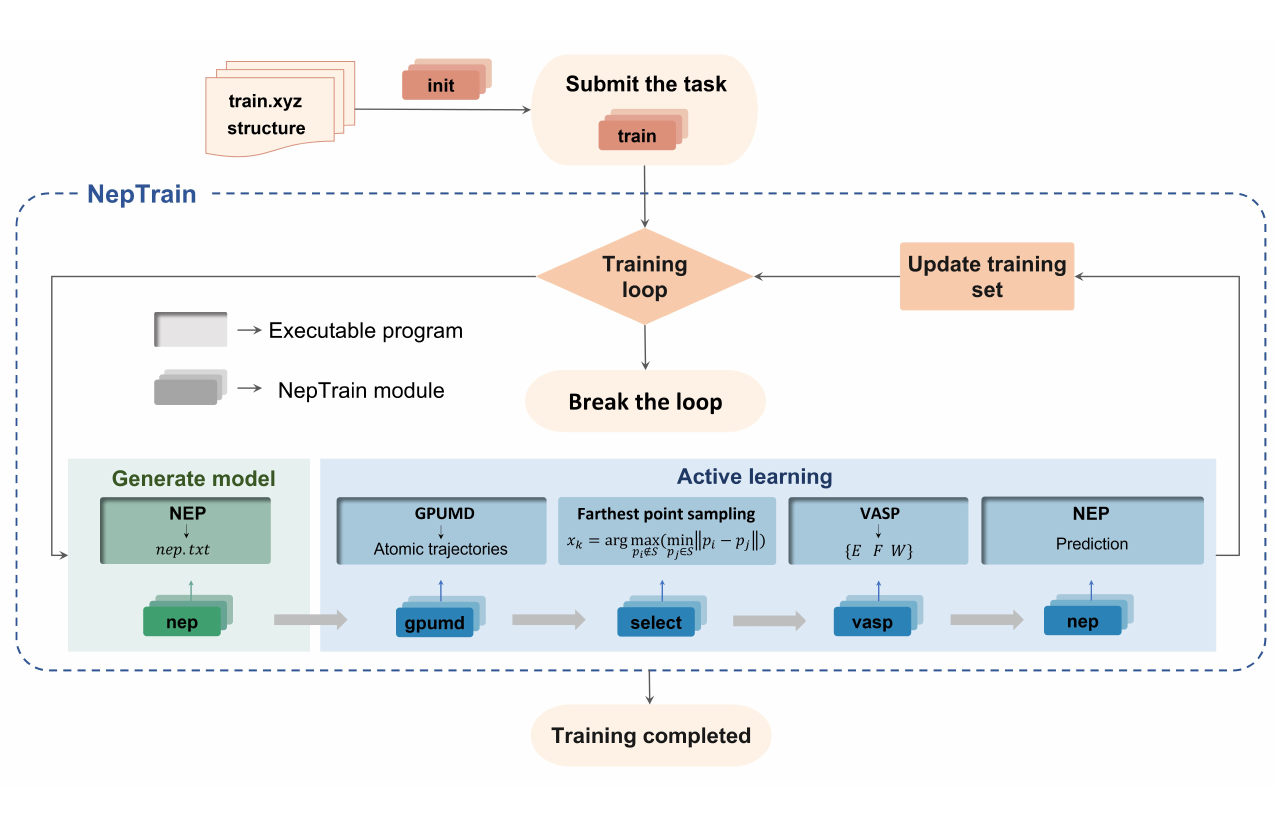} 
    \caption{Workflow of the automated training process in NepTrain. Starting from the prepared \texttt{train.xyz} and structure files, the \texttt{init} module initializes the task, and the \texttt{train} module sequentially calls the core modules:  \texttt{nep}, \texttt{gpumd}, \texttt{select}, and \texttt{vasp}. The prediction for the \gls{nep} model is carried out within the \texttt{nep} module. In the figure, E represents energy, F denotes force, and W stands for virial. Simple training datasets can be generated using NepTrain’s \texttt{perturb} module, while more advanced dataset preparation is supported through NepTrainkit.}
    \label{fig:neptrain_framework}
\end{figure*}

\section{NepTrain}
\label{sec:neptrain}

NepTrain invokes tools such as \gls{nep}, \gls{gpumd} and \gls{vasp} to facilitate the automated training of \gls{nep} models. It utilizes the Python package ASE \cite{Hjorthlarsen2017TheAtomic} for reading and writing lattice structures. The software operates using the format \texttt{NepTrain command argv}, where \texttt{command} specifies the functional module, and \texttt{argv} provides the corresponding parameters for that module. Furthermore, all commands support execution in both slurm environments and local setups. Users can access relevant help information by running \texttt{NepTrain -h} or \texttt{NepTrain command -h}. For detailed installation instructions and a comprehensive user guide, please refer to the documentation: \url{https://neptrain.readthedocs.io/en/latest/}.

The overall workflow of the NepTrain automation training framework is illustrated in \autoref{fig:neptrain_framework}. The framework is designed to efficiently assist users in training \gls{nep} models, and can be broadly divided into two main phases: initialization and training loop. In the initialization phase, the system first reads the user-configured yaml file to load the necessary training parameters. Next, based on the configuration file, the task execution mode (local or slurm cluster) is selected, and the task manager is initialized. Once the training loop begins, the system uses the train module to sequentially invoke and execute the \texttt{nep}, \texttt{gpumd}, \texttt{select}, and \texttt{vasp} modules. The training set is updated based on the selected data and predicted results, and the next iteration is initiated. This process continues until the maximum iteration limit is reached, at which point the system exits the training loop and completes the \gls{nep} model training.

In the following, we will provide a detailed introduction to the functionality of each module. This section focuses on the working mechanisms of the program, while specific application cases will be elaborated in Section \ref{sec:example}.

 \subsection{The \texttt{perturb} module}

This module is designed for perturbing the training set. In order to increase the diversity of the training set and the stability of the \gls{nep} model, we provide two functions: cell perturbation and atomic position perturbation. Users can invoke this functionality by executing the command: \texttt{NepTrain perturb structure\_path [option]}. Specifically, the \texttt{-n} parameter specifies the number of perturbed structures to generate, while the \texttt{-c} and \texttt{-d} parameters control the magnitude of the cell and atomic position perturbations, respectively.
To generate high-quality perturbed structures, we recommend the following steps: First, users can generate 10000 perturbed structures with the following command: \texttt{NepTrain perturb xxx.vasp -n 10000}. The perturbed structure file, \texttt{perturb.xyz}, will be saved in the current working directory. Next, users can use the select command to choose 100 structures from this set as the training set: \texttt{NepTrain select perturb.xyz -max 100}. The selected structures will be saved in the \texttt{selected.xyz} file, and the selection results will be visualized in the \texttt{selected.png} file. Using this approach, users can ensure that the perturbed structures comprehensively cover the perturbation dataset, providing a diverse supplement to the training set.

\subsection{The \texttt{nep} module}

This module is designed to invoke the executable program \texttt{nep} for potential training. If a training dataset file named \texttt{train.xyz} is present in the current directory, the training task can be initiated directly by executing the command \texttt{NepTrain nep}. For training dataset files with different names, users can specify the file path using the command \texttt{NepTrain nep -train path}. By default, the program automatically infers the elemental composition from the training dataset and generates a \texttt{nep.in} file. Alternatively, users can provide a custom \texttt{nep.in} file using the \texttt{-in nep.in} parameter. Furthermore, predictions can be performed using the \texttt{-pred} option.
After each training iteration, the program automatically generates data plots illustrating the training errors of energy, forces, stress, and virial. These plots will be saved in PNG format in the working directory .

\subsection{The \texttt{gpumd} module}
This module is responsible for performing \gls{md} simulations. Users can initiate a simulation by executing the command \texttt{NepTrain gpumd structure\_path }, where \texttt{structure\_path} is a required parameter specifying the path to the structure file. The input file for \gls{gpumd} is expected to be \texttt{run.in} in the current directory. If this file is absent, the program will automatically generate a default input file configured for the NPT ensemble. Users can customize the simulation time (in ps, default: 10 ps) using the \texttt{--time} parameter and set the simulation temperature (default: 300 K) with the \texttt{--temperature} parameter. Upon completion of the \gls{md} simulation, the program automatically generates a time-energy line plot, which is saved in PNG format in the working directory .

\subsection{The \texttt{vasp} module}

In this module, we selected \gls{vasp} as the tool for single-point energy calculations due to its excellent performance in terms of convergence and compatibility. Similar to the \texttt{gpumd} module, the \texttt{vasp} module also requires a structure path to be specified. Users can perform parallel computations in a 64-core computational environment with the following command: \texttt{NepTrain vasp structure\_path -np 64}. This module calls \gls{vasp} via the ASE wrapper, with the Monkhorst-Pack grid being the default k-points type. To enhance compatibility, we have designed two mutually exclusive k-points input options: \texttt{--kspacing} and \texttt{--ka}. Upon completion of the calculation, the program automatically extracts the output file in XYZ format. Users can specify the output path via the \texttt{--out} parameter. This design aims to increase the flexibility of the computational process while simplifying user operations, making it applicable across different computational environments.
 
\subsection{The \texttt{select} module}
\label{subsec:select_module}

During the iterative process, selecting suitable structures from the \gls{md} trajectory to expand the training set is a crucial step \cite{Jager2018Machine,Bartok2017Machine}. We employ the farthest-point sampling method to select descriptors for all structures, thereby identifying the necessary structures from the raw dataset trajectory to expand the training set. The basic principle of farthest-point sampling is as follows:
\begin{enumerate}
    \item Initialize the selected point set \( S = \{p_1\} \), where \( p_1 \) is an arbitrary starting point.
    \item For each unselected point \( p_i \), calculate its shortest distance to the selected point set \( S \):
    \[
    d(p_i, S) = \min_{p_j \in S} \|p_i - p_j\|
    \]
    \item Select the point \( p_k \) that maximizes \( d(p_i, S) \):
    \[
    p_k = \arg\max_{p_i \notin S} d(p_i, S)
    \]
    \item Add \( p_k \) to the selected point set \( S \).
    \item Repeat steps 2-4 until the size of \( S \) reaches the desired number of points.
\end{enumerate}
Meanwhile, the \texttt{-f} or \texttt{---filter} parameter allows users to enable or disable filtering based on the minimum bond length (disabled by default). This functionality helps to exclude non-physical structures from the \gls{md} trajectory. The underlying principle of this filtering method is detailed in Section \ref{subsec:Dataset Manipulation}. If a potential is available, the program will obtain the descriptors via \texttt{NEP\_CPU}; otherwise, the Python package DScribe is used to generate \gls{soap} descriptors \cite{Himanen2020DScribe}. Additionally, the program supports dimensionality reduction using \gls{pca} \cite{Pedregosa2011Scikitlearn} and \gls{umap} \cite{Mcinnes2020UMAP} methods, and can generate corresponding plots. Users can invoke this functionality by executing the following command: \texttt{NepTrain select *trajectory\_path -base train.xyz option}.

\subsection{The \texttt{train} module}

The train module is the core component of the NepTrain software, responsible for integrating and coordinating the execution of submodules such as \texttt{nep}, \texttt{gpumd}, \texttt{select}, and \texttt{vasp}. Users can initialize the task template by executing the \texttt{NepTrain init} command, which generates necessary files, including \texttt{job.yaml}, \texttt{run.in}, structure files, and submission scripts.  The \texttt{job.yaml} file is used to control training parameters, including the root path of the working directory (\texttt{work\_path}, default is \texttt{./cache}), the starting task (\texttt{current\_job}, which can be set to \texttt{vasp}, \texttt{nep}, or \texttt{gpumd}), the list of MD simulation time steps (\texttt{step\_times}, which determines the maximum number of iterations), and the temperature settings for each step (\texttt{temperature\_every\_step}), among others. Most of the default parameters are highly generalizable. Detailed parameter information can be found in the documentation.
To facilitate the management of output files, all computational details are saved in the \texttt{work\_path} directory, with the detailed files for each iteration stored in a folder named \texttt{Generation-*}. After completing the necessary configurations, users can start the training task by executing the following command: \texttt{NepTrain train job.yaml}. This module automates the execution of cumbersome steps by integrating various computational tools, significantly reducing manual intervention and allowing researchers to focus more on result analysis and parameter optimization.

\section{NepTrainKit}
\label{sec:neptrainkit}

Prior to this work, researchers primarily used OVITO \cite{Stukowski2010Visualization} or VMD \cite{William1996VMD} to inspect dataset structures and employed Python scripts to visualize \gls{nep} output. However, these two tools operate independently, lacking the capability for integrated analysis. In contrast, Chemiscope enables users to intuitively explore structural features by projecting structure descriptors onto a two-dimensional coordinate system \cite{fraux2020Chemiscope}. Against this backdrop, we designed and developed NepTrainKit, which integrates \gls{nep} training datasets with property visualization, enabling seamless interactive analysis.

\begin{figure*}[htbp]
    \centering
    \includegraphics[width=0.98\textwidth]{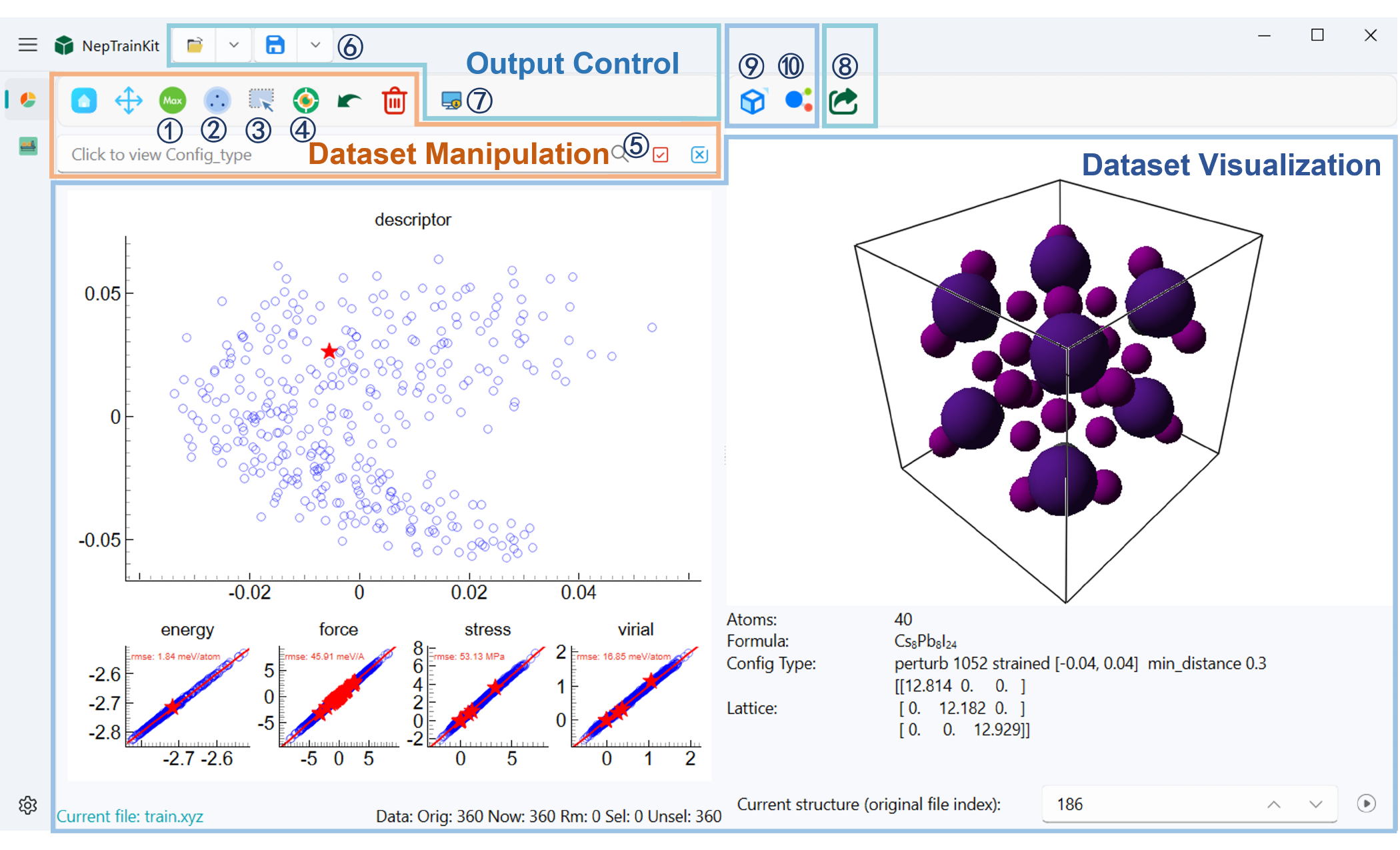} 
    \caption{Interface of NepTrainKit, illustrating the three primary sections: Dataset Manipulation(1. Maximum error point selection; 2. Farthest point sampling; 3. Manual selection; 4. Non-physical structure detection; 5. Config\_type keyword configuration), Dataset Visualization (9. Structural view switching; 10. Atom display mode switching), and Output Control (6. Data import and export; 7. Descriptor export; 8. Standard xyz file export). This toolkit is designed to  simplify and optimize the \gls{nep} model training process, providing an intuitive graphical interface and analysis tools.}
    \label{fig:neptrainkit_framework1}
\end{figure*}

NepTrainKit is a toolkit focused on the manipulation and visualization of \gls{nep} training datasets, providing an intuitive graphical interface and analytical tools to help users efficiently refine their training sets. The software is developed in Python due to its well-established ecosystem, which offers a wide range of built-in modules, allowing us to focus on business logic development rather than low-level implementation. Given that the program primarily runs on \gls{hpc} platforms, we chose PySide6 as the cross-platform GUI framework. For the rendering engine, NepTrainKit supports both pyqtgraph and vispy\cite{Campagnola2025Vispy}, allowing users to freely switch between them in the settings. Each module has its own strengths and use cases: pyqtgraph is well-suited for small to medium datasets ($<10^5$ scatter points) and offers strong cross-platform compatibility, though it may struggle with extremely large datasets; vispy, on the other hand, leverages OpenGL for high-performance rendering, maintaining smooth visualization even for datasets as large as $10^6$ scatter points, but requires GPU support. To enhance usability, NepTrainKit has been compiled into a standalone executable for Windows and is also available via pip, significantly improving cross-platform compatibility. It supports Ubuntu, Windows, and macOS operating systems. For detailed usage documentation, please refer to: \url{https://neptrainkit.readthedocs.io/en/latest}

The functional interface of NepTrainKit, as shown in \autoref{fig:neptrainkit_framework1}, is divided into three main sections: Dataset Visualization, Dataset Manipulation, and Output Control. The software first reads \gls{nep} output files from the current working directory, including mandatory \texttt{*.xyz} files and optional files such as \texttt{nep.txt}, \texttt{energy\_*.out}, \texttt{force\_*.out}, \texttt{stress\_*.out}, \texttt{virial\_*.out}, \texttt{dipole\_*.out}, and \texttt{polarizability\_*.out}. These files serve as the basis for \gls{nep} training dataset manipulation and visualization. NepTrainKit is equipped with a pre-trained \gls{nep} model spanning 89 chemical elements, serving as a general-purpose default potential.  If the directory contains only a \texttt{*.xyz} file, NepTrainKit will automatically invoke \texttt{NEP\_CPU} to compute the properties of each frame in the file and store the results in the working directory. This caching mechanism accelerates future loading processes, enhancing overall efficiency and usability.

\subsection{Dataset visualization}
In the dataset visualization module, we provide graphical representations of various data types, including descriptors, energy, force, stress, virial stress, dipole moment, and polarizability. For descriptor processing, we first use \texttt{NEP\_CPU} to extract atomic features and compute their average values to generate structural descriptors. These descriptors are then projected into a two-dimensional space using \gls{pca} to facilitate dataset diversity analysis, such as distinguishing structural clusters. 

Additionally, this module supports frame-by-frame visualization of crystal structures, allowing users to intuitively observe the dynamic changes in the structures along with relevant structural information. During crystal structure visualization, atomic sizes are determined by their covalent radii, and colors follow the CPK color scheme, ensuring effective visual representation and clear information conveyance. Users can also choose between a ball-and-stick model and a space-filling atomic model, as well as switch between perspective and orthographic viewing modes, to accommodate different visualization preferences and inspection needs.

To assist in assessing the physical validity of structures, the system includes a minimum atomic pair bond length display function. The software calculates all pairwise atomic distances under periodic boundary conditions for the selected structure and highlights excessively short distances in red, making it easier to identify potentially non-physical structures. Furthermore, within the structure display area, all atomic pairs that fail to meet logical bond length criteria are highlighted, enhancing users’ intuitive perception of structural anomalies.

\subsection{Dataset Manipulation}
\label{subsec:Dataset Manipulation}
In the data processing module, we have integrated four key tools: Maximum Error Point Selection, Farthest Point Sampling, Manual Selection, and Non-physical Structure Detection. Users can interactively mark or remove scatter points and their corresponding structures using mouse operations. Additionally, \texttt{Config\_type} keywords can be combined with the toolbar to enable more advanced selection logic. When the status of a structure changes, the corresponding scatter points are highlighted in red across all four subplots. For instance, if certain structures are selected in the energy distribution plot, they will be simultaneously highlighted in other subplots. This feature provides an intuitive representation of a structure’s position and influence within the dataset, facilitating efficient training set adjustments. The following sections provide a detailed introduction to these four tools and the \texttt{Config\_type} keyword selection functionality.

\textbf{Maximum Error Point Selection}: This tool identifies the $N$ structures with the largest errors, allowing users to quickly locate and remove anomalous data. Users can specify an integer $N$, and the program will select the top $N$ structures with the largest errors based on the current main plot, marking them for easier data inspection and cleaning. The selection is based on the following principle:
For each structure $i$, the deviation between the \gls{nep}-predicted structure descriptor $D_i^{\text{NEP}}$ and the \gls{dft} reference descriptor $D_i^{\text{DFT}}$ is computed. The sum of the absolute errors across all descriptor dimensions is then calculated:
\begin{equation}
E_i = \sum_{j=0}^{C-1} \left( D_{i,j}^{\text{NEP}} - D_{i,j}^{\text{DFT}} \right), \quad i = 0, 1, \dots, N-1
\end{equation}
Structures with larger errors may indicate anomalies and require further inspection or removal.

\textbf{Farthest point sampling}: This tool performs farthest point sampling on descriptor data based on a user-specified maximum retention number and minimum distance. By selecting representative structures, it sparsifies the dataset while preserving sample diversity as much as possible. In NepTrainKit, farthest point sampling is applied based on structural descriptors, ensuring that retained structures are well-distributed in descriptor space. The implementation details can be found in Section \ref{subsec:select_module}. Notably, to enhance user convenience, the program automatically applies an inverse selection operation after the sampling process. Specifically, structures not selected by farthest point sampling will be highlighted in red within the software interface, allowing users to delete them with a single click.

\textbf{Manual selection}: This tool provides a flexible way to manually mark structures. When activated, the main plot enters edit mode, allowing users to select structures by clicking individual points or drawing a selection region with the left mouse button. To deselect a structure, users can simply right-click on the corresponding point.

\textbf{Non-physical structure detection}: This tool automatically identifies non-physical structures based on bond length thresholds. The core determination logic is as follows:
\begin{equation}
(R_1 + R_2) \times \text{Coeff} > \text{bond}
\end{equation}
where $\rm R_1$ and $\rm R_2$ represent the covalent radii of the atomic pair, bond is the bond length considering lattice periodicity, and Coeff is a covalent radius coefficient, which can be adjusted in the settings according to the specific system. The program iterates through all atomic pairs in the training set. If the above condition is met, the atomic pair is considered to have a non-physical bond length, and the entire structure is flagged as a non-physical structure.

\textbf{Config\_type keyword selection}: This tool allows users to filter structures by selecting a specific configuration type and searching using prefixes, suffixes, or keywords. 
The labels used for filtering are extracted from the comment lines of the xyz format files. After entering the search criteria and clicking the Search button, all matching structures will be highlighted in green on the main plot to indicate the search results. It is important to note that green highlighting only indicates a match and does not automatically select the structures. If further actions are needed, users can manually select or deselect the highlighted structures by clicking the Select (Deselect) button.

By utilizing these tools individually or in combination, users can efficiently and conveniently process datasets. The visualization of the training set not only helps eliminate noisy data, minimizing its impact on model training, but also provides an intuitive view of the dataset's distribution. This, in turn, offers valuable insights for subsequent model optimization and parameter adjustments.

\subsection{Output control}
The Output control module is designed to manage the output files generated during the \gls{nep} dataset visualization process, ensuring convenient data storage and subsequent analysis. This module supports automatic generation and saving of output files, including dataset visualization images and data files containing descriptors, energy, force, stress, and virial information. 

Additionally, the software includes a structure descriptor export tool, which allows users to select the desired structure type using a mouse tool and export the corresponding structure descriptors to a specified path. This feature facilitates further analysis and the creation of structural data distribution plots, providing a more intuitive view of the training data distribution. A detailed example of this will be presented in Section \ref{sec:example}. Users can also export the structure of the current frame as a standard \texttt{*.xyz} file for subsequent structural analysis and computational tasks. By offering comprehensive output options, this module enhances the flexibility and practicality of NepTrainKit, enabling users to efficiently manage, store, and analyze output data.

\section{Examples}
\label{sec:example}

In this section, we present the details of a fully automated workflow for generating a \gls{nep} model, using $\rm CsPbI_3$ as a representative example and implementing the process through NepTrain. We demonstrate that this tool enables the construction of highly accurate \gls{nep} models, even when trained on a small dataset, and facilitates the reliable prediction of a wide range of material properties.

\subsection{Constructing the NEP model}

\textbf{Training Set}: $\rm CsPbI_3$ is a prototypical inorganic halide perovskite, which has achieved a record power conversion efficiency (PCE) of 21.24\% in perovskite solar cells \cite{Xu2025Recordefficiency}. 
It typically exists in three polymorphs: the $\alpha$-phase (\textit{Pm-3m}, 645 K), the $\beta$-phase (\textit{P4/mbm}, 510 K), and the $\gamma$-phase (\textit{Pbnm}, 325 K) \cite{Marronnier2018Anharmonicity}. 
Accordingly, we constructed a perturbation dataset comprising these three phases. 
Initially, we used the VESTA software \cite{momma2011VESTA3} to construct supercells with atom counts ranging between 40 and 160 for the three structural types of $\rm CsPbI_3$. 
These supercells served as the initial structures for the training set. 
The \texttt{structure} folder contains three structure files corresponding to different cell sizes. 
Subsequently, a perturbation dataset was generated by applying a 4\% strain to the unit cell and a 0.3 Å atomic coordination perturbation. 
To avoid generating non-physical structures induced by perturbations, we enabled bond-length filtering using the \texttt{-f} parameter: \texttt{NepTrain perturb structure -n 5000 -c 0.04 -d 0.3 -f}.
We selected 200 representative structures from the dataset with 0.001 Å as the maximum atomic perturbation distance, and completed the training set preparation by renaming \texttt{selected.xyz} to \texttt{train.xyz}: \texttt{NepTrain select perturb.xyz -max 200 -d 0.001}; \texttt{mv selected.xyz train.xyz}

\textbf{DFT calculations}: To obtain accurate single-point energy data, we performed calculations using \gls{vasp} for all the selected structures \cite{Kresse1994AbInitio,Kresse1996Efficient}. The exchange-correlation functional with the \gls{pbe} form is applied for calculations \cite{Blochl1994Projector,Perdew1996Generalized}. The cutoff energy is set to be 500 eV. The Brillouin zone is sampled by the $\Gamma$-centered k-point mesh, with reciprocal lattice spacing less than 0.2 Å$\rm ^{-1}$. The convergence criterion for the total energy in the self-consistent structural calculations is set to be less than $\rm 10^{-6}$ eV.

\textbf{Active learning}: The active learning process is fully automated and incorporates a bond-length filtering method implemented using NepTrain. Task management and execution are automated through the use of specific configuration files. Specifically, the system automatically sets the \gls{md} simulation time steps and temperature ranges for different stages using the \texttt{step\_times} and \texttt{temperature\_every\_step parameters}, ensuring comprehensive coverage of various temperature conditions. In this case, \gls{md} simulations are conducted under the NPT ensemble using the stochastic cell rescaling method \cite{Bernetti2020Pressure}. The simulation temperature ranges from 0 K to 620 K, with iteration durations set to 100 ps, 500 ps, 1 ns, and 5 ns, respectively. This configuration allows the system to automatically adjust both temperature and time step during the active learning process, enabling thorough structural sampling and thereby improving the quality and accuracy of model training. The detailed configuration file (\texttt{job.yaml}) used for running these simulations is provided in the online repository (see \textit{Data Availability Statement} for details).

\subsection{Validating the NEP Model}
\begin{figure}[htbp]
    \centering
    \includegraphics[width=0.5\textwidth]{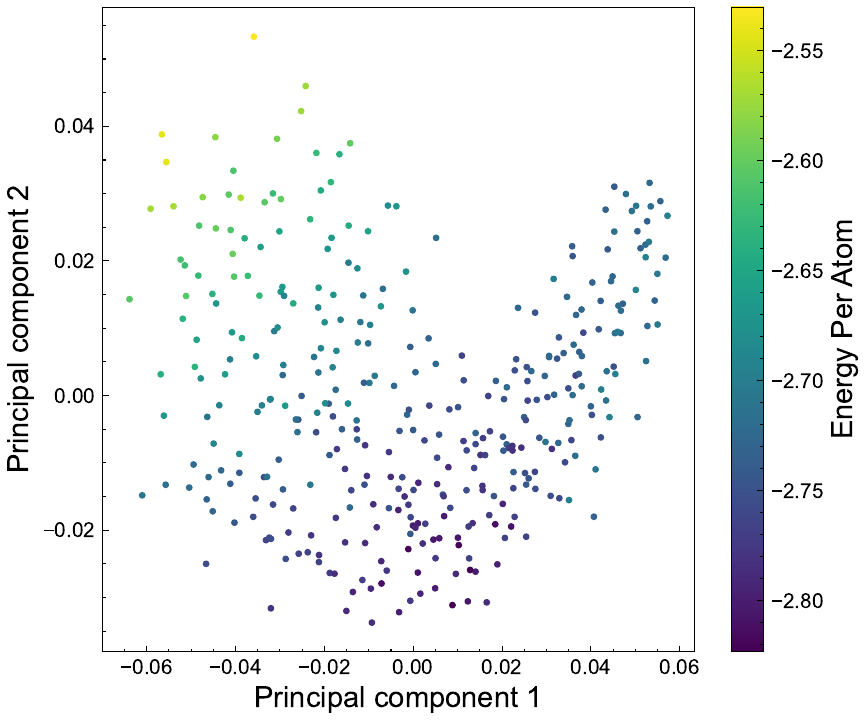} 
    \caption{The projected schematic of structural descriptors for $\rm CsPbI_3$.}
    \label{fig:descriptors}
\end{figure}

First, we plotted a projection diagram of the structural descriptors for $\rm CsPbI_3$ to assess both the diversity of the dataset and the quality of the descriptors, as shown in \autoref{fig:descriptors}. Each point represents an independent structure, and its spatial distribution is determined by dimensionality reduction of the structure descriptors via principal component analysis (PCA). The color of each point indicates the energy of the structure. 
The data points are uniformly distributed with minimal overlap, indicating that, although the training set consists of only 200 representative structures, it effectively spans a wide range of structural configurations.

\begin{figure}[htbp]
    \centering
    \includegraphics[width=0.5\textwidth]{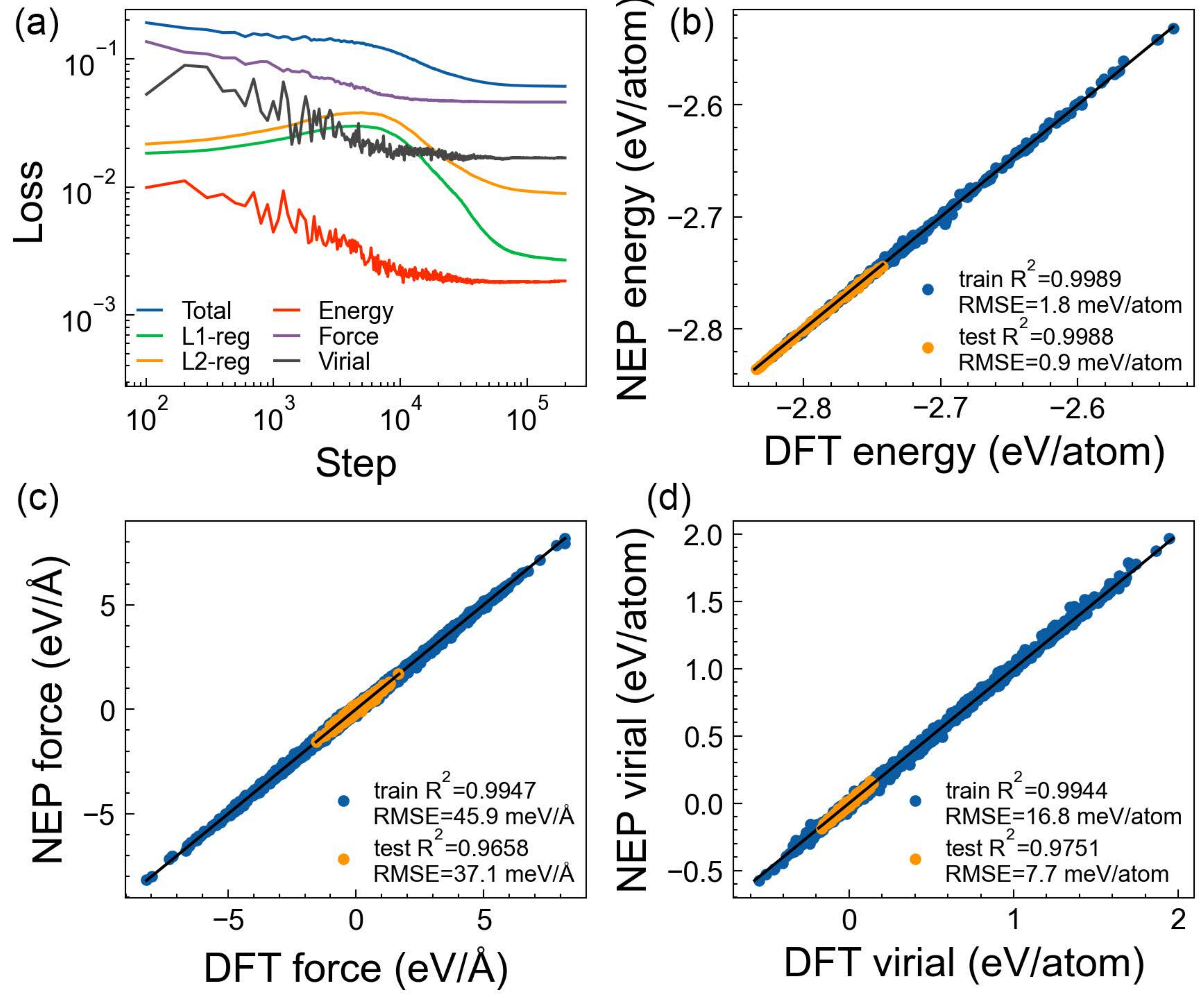} 
    \caption{ (a) Evolution of
the various loss terms as a function of the training step. Comparison of (b) energy, (c) force, and (d) virial calculated using the \gls{nep} model with DFT reference data.}
    \label{fig:loss_function}
\end{figure}

Then, we evaluated the fitting performance of the \gls{nep} model with respect to  energy, force, and virial. As shown in \autoref{fig:loss_function}, the \gls{nep} model demonstrates high accuracy on both the training and test datasets, achieving \gls{rmse} values for energy, force, and virial of less than 1.9 meV/atom, 46.0 meV/Å, and 17.0 meV/atom, respectively. Specifically, to construct the test dataset, we first performed \gls{md} simulations using \gls{gpumd} for the orthorhombic phase of $\rm CsPbI_3$ (containing 80 atoms) over a temperature range of 0–600 K, with a total simulation time of 10 ns. We then uniformly sampled the resulting trajectory, selecting 100 representative structures to serve as the test set. Notably, the \gls{rmse} values for energy and virial in the test dataset are approximately half of those observed in the training dataset.  These performance metrics are also comparable to those reported in the literature for a $\rm CsPbI_3$  \gls{nep} model constructed using the same PBE functional and 510 structures, which exhibited \gls{rmse}s of 0.7 meV/atom, 43.7 meV/Å, and 9.6 meV/atom for energy, force, and virial, respectively \cite{Wiktor2023Quantifying}. This result indicates that the \gls{nep} potential for $\rm CsPbI_3$ , generated from only 200 perturbed configurations and active learing, can still achieve acceptable accuracy.

\begin{figure}[htbp]
    \centering
    \includegraphics[width=0.5\textwidth]{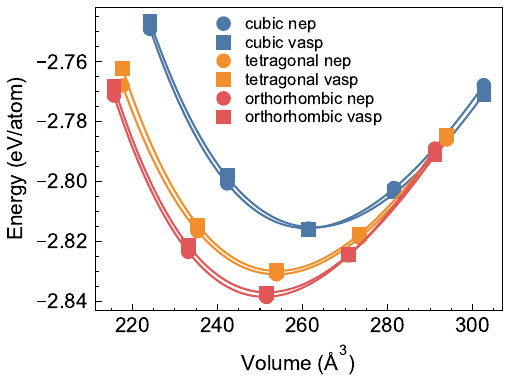} 
    \caption{ EOS curves for $\rm CsPbI_3$. Circular and square markers represent the calculation results from \gls{nep} and \gls{vasp}, respectively. The cubic, tetragonal, and orthorhombic phases of $\rm CsPbI_3$ are indicated by blue, orange, and red curves, respectively.  }
    \label{fig:EOS}
\end{figure}

\begin{figure*}[tb]
    \centering
    \includegraphics[width=0.98\textwidth]{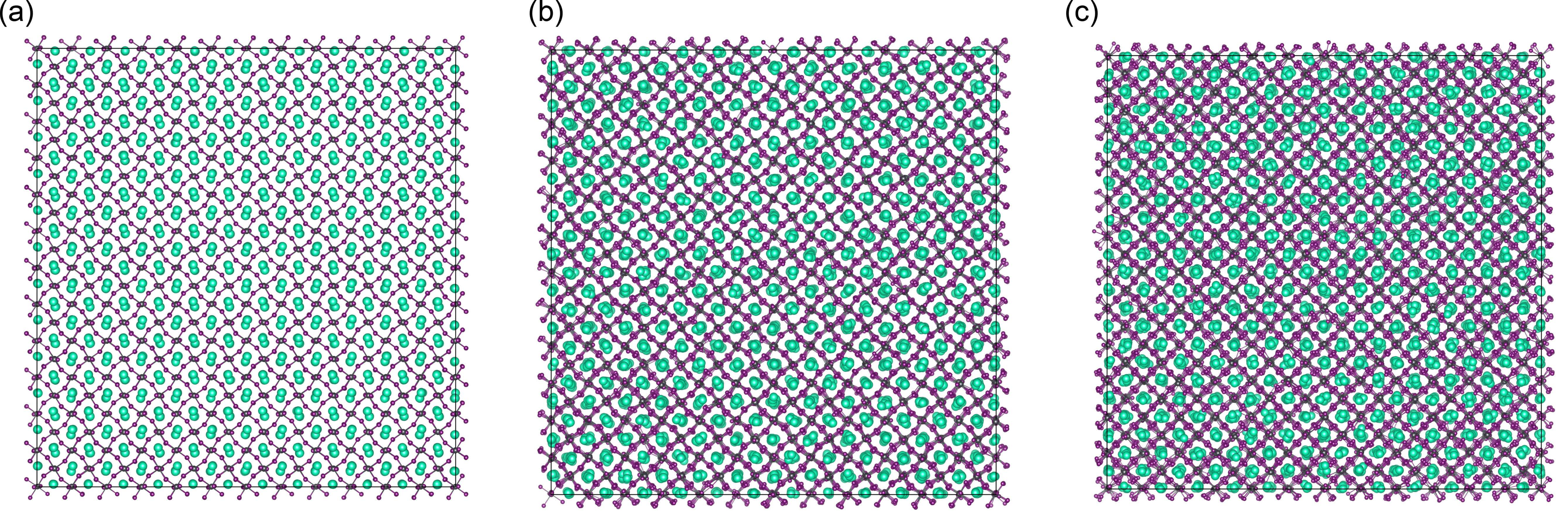} 
    \caption{Schematic illustration of the evolution of the $\rm CsPbI_3$ supercell structure. (a) Orthorhombic phase atomic structure at 20 K. (b) Tetragonal phase atomic structure upon heating from 20 K to 320 K. (c) Cubic phase atomic structure upon heating from 320 K to 500 K.}
    \label{fig:crystal_structure}
\end{figure*}

Finally, we tested the performance of the \gls{nep} model for $\rm CsPbI_3$ on several properties. The predictions for the \gls{eos} obtained from both \gls{dft} and \gls{nep} are shown in \autoref{fig:EOS}. It is evident that the \gls{nep} model reproduces well the \gls{dft} results for the cubic, tetragonal, and orthorhombic phases of $\rm CsPbI_3$ considered in this study. Further, we applied \gls{gpumd} simulations to investigate the phase transition of $\rm CsPbI_3$. To eliminate size effects, we used a supercell comprising 23040 atoms, equivalent to $12\times12\times8$ primitive orthorhombic unit cells. The crystal structures of $\rm CsPbI_3$ in different phases are shown in \autoref{fig:crystal_structure}, where (a)-(c) correspond to the orthorhombic phase at 20 K, the tetragonal phase at 320 K and the cubic phase at 500 K, respectively. During the simulation, the system is first relaxed under the NPT ensemble at the initial temperature and zero external pressure for 0.1 ns with a time step of 1 fs. The total simulation time was set to 5 ns, with the temperature increased from 20 K to 620 K in intervals of 100 K, corresponding to a heating rate of 120 K/ns. To facilitate the analysis of phase transitions among the three phases of $\rm CsPbI_3$, we defined the normalized lattice constants. For example, since the orthorhombic phase contains four formula units, the normalized lattice constants were set as $a=a_0/\sqrt{2}$, $b=b_0/\sqrt{2}$, $c=c_0/2$, where $a_0$, $b_0$ and $c_0$ are lattice constants of the unit cells.

\begin{figure}[htbp]
    \centering
    \includegraphics[width=0.5\textwidth]{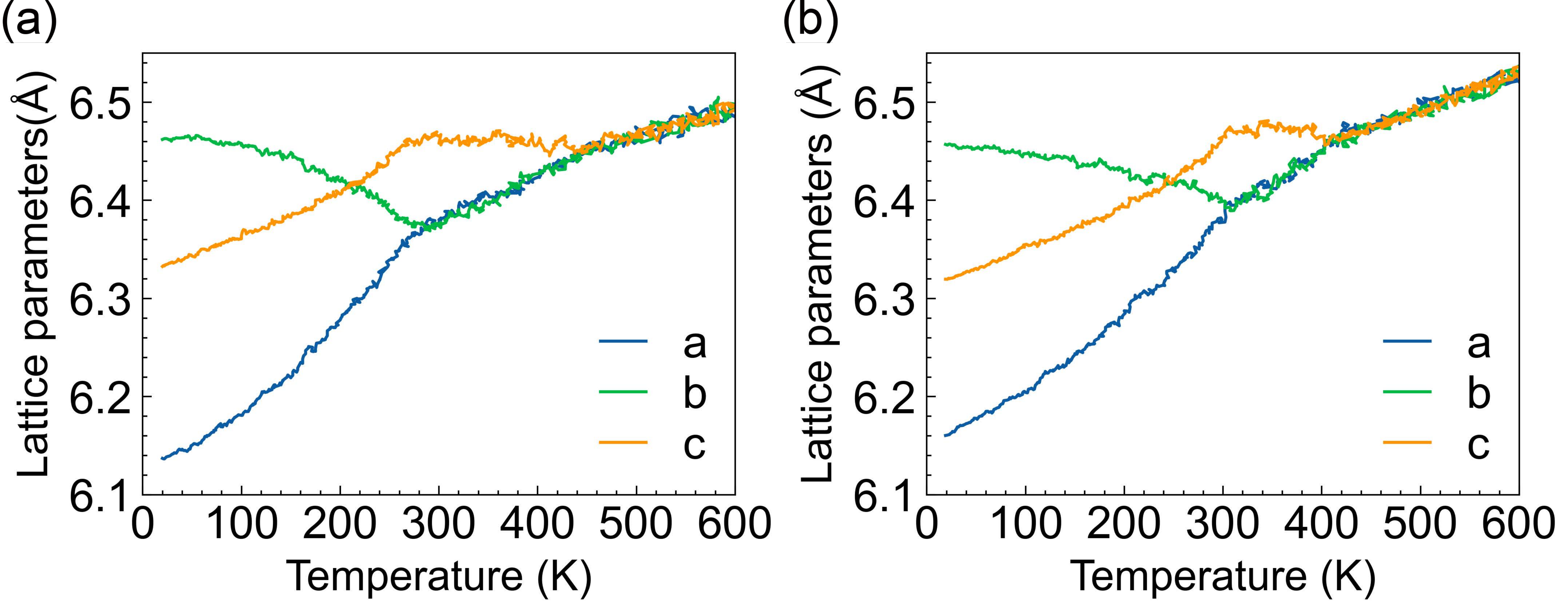} 
    \caption{Lattice parameters of $\rm CsPbI_3$ as a function of temperature obtained from MD simulations during the heating process: (a) results from the \gls{nep} model reported in the reference \cite{Fransson2023Phase}; (b) results from the \gls{nep} model developed in this study.}
    \label{fig:lattice_phase}
\end{figure}

The evolution of the normalized lattice constants as a function of temperature is shown in \autoref{fig:lattice_phase}. As the temperature increases, the lattice constants $a$ and $c$ gradually increase, while $b$ decreases. At $a$ = $b$ , the system undergoes a transition from the orthorhombic phase to tetragonal phase, with the transition temperature approximately at 309 K. Then, as we further increase the temperature, $a$ and $b$ increase while $c$ decreases. Finally, the normalized lattice constants of $a$, $b$ and $c$ reach the same value, signaling a phase transition from the tetragonal to the cubic phase, with the transition temperature around 404 K. Furthermore, using the results reported by Fransson et al.\cite{Fransson2023Phase} as a reference (with phase transition temperatures of 297 K and 438 K), we observed good agreement with our calculated temperature values. Therefore, through simulations of the \gls{eos} and phase transition processes, we confirm that employing NepTrain to generate training sets containing only 200 perturbed configurations, combined with automated iterative training, yields an \gls{nep} model for $\rm CsPbI_3$ with acceptable accuracy. This workflow can also be readily applied to other material systems.

\section{Summary and conclusions}

In this work, we introduced the software tools NepTrain and NepTrainKit for the convenient preparation and manipulation of training datasets to construct high-quality training sets for \gls{nep} models. NepTrain integrates various functionalities, including dataset generation, model training, and active learning, by leveraging tools such as \gls{gpumd}, \gls{vasp}, and \gls{nep}. By combining Python scripts and Bash commands, NepTrain enables automated training while maintaining user flexibility in operations. Additionally, NepTrain incorporates mechanisms such as minimum bond-length filtering and farthest-point sampling, which effectively filter out non-physical structures and enhance the representativeness of the training set. These features ensure the efficient training of highly accurate \gls{nep} models. NepTrainKit provides an intuitive graphical interface with excellent cross-platform compatibility, allowing users to efficiently manage and optimize training datasets.

As a case study demonstrating the capabilities of this toolkit, we used NepTrain to develop a $\rm CsPbI_3$ \gls{nep} model and conducted performance tests. The results highlight the advantages of NepTrain, including its low requirement for the size of the training dataset, its automated training workflow, and the high quality of the resulting \gls{nep} models. Most importantly, NepTrain enables users to achieve fully automated training with high efficiency and minimal resource consumption, offering an optimal balance between accuracy and computational cost. Furthermore, the high-quality training datasets constructed using NepTrain and NepTrainKit can also be extended to the training of other machine learning interatomic potential (MLIP) frameworks beyond \gls{nep}, thereby expanding their applicability within the field.

\vspace{0.5cm}

\begin{acknowledgments}
C.C. and Z.W. were supported by the Natural Science Foundation of Guangxi province (No. 2024GXNSFAA010428). G. T. was supported by Beijing Institute of Technology Research Fund Program for Young Scholars (Grant No. XSQD-202222008) and Guangdong Key Laboratory of Electronic Functional Materials and Devices Open Fund (EFMD2023004M). 
\end{acknowledgments}

\vspace{0.5cm}
\noindent{\textbf{CONFLICT OF INTEREST STATEMENT}}

The authors have no conflicts to disclose. 

\vspace{0.5cm}
\noindent{\textbf{DATA AVAILABILITY STATEMENT}}

The NepTrain package, including data for the $\rm CsPbI_3$ example, is freely available at \url{https://github.com/aboys-cb/NepTrain}.
The NepTrainKit package is freely available at \url{https://github.com/aboys-cb/NepTrainKit}. 


\bibliography{refs}

\begin{thebibliography}{56}%
\makeatletter
\providecommand \@ifxundefined [1]{%
 \@ifx{#1\undefined}
}%
\providecommand \@ifnum [1]{%
 \ifnum #1\expandafter \@firstoftwo
 \else \expandafter \@secondoftwo
 \fi
}%
\providecommand \@ifx [1]{%
 \ifx #1\expandafter \@firstoftwo
 \else \expandafter \@secondoftwo
 \fi
}%
\providecommand \natexlab [1]{#1}%
\providecommand \enquote  [1]{``#1''}%
\providecommand \bibnamefont  [1]{#1}%
\providecommand \bibfnamefont [1]{#1}%
\providecommand \citenamefont [1]{#1}%
\providecommand \href@noop [0]{\@secondoftwo}%
\providecommand \href [0]{\begingroup \@sanitize@url \@href}%
\providecommand \@href[1]{\@@startlink{#1}\@@href}%
\providecommand \@@href[1]{\endgroup#1\@@endlink}%
\providecommand \@sanitize@url [0]{\catcode `\\12\catcode `\$12\catcode `\&12\catcode `\#12\catcode `\^12\catcode `\_12\catcode `\%12\relax}%
\providecommand \@@startlink[1]{}%
\providecommand \@@endlink[0]{}%
\providecommand \url  [0]{\begingroup\@sanitize@url \@url }%
\providecommand \@url [1]{\endgroup\@href {#1}{\urlprefix }}%
\providecommand \urlprefix  [0]{URL }%
\providecommand \Eprint [0]{\href }%
\providecommand \doibase [0]{https://doi.org/}%
\providecommand \selectlanguage [0]{\@gobble}%
\providecommand \bibinfo  [0]{\@secondoftwo}%
\providecommand \bibfield  [0]{\@secondoftwo}%
\providecommand \translation [1]{[#1]}%
\providecommand \BibitemOpen [0]{}%
\providecommand \bibitemStop [0]{}%
\providecommand \bibitemNoStop [0]{.\EOS\space}%
\providecommand \EOS [0]{\spacefactor3000\relax}%
\providecommand \BibitemShut  [1]{\csname bibitem#1\endcsname}%
\let\auto@bib@innerbib\@empty
\bibitem [{\citenamefont {Unke}\ \emph {et~al.}(2021)\citenamefont {Unke}, \citenamefont {Chmiela}, \citenamefont {Sauceda}, \citenamefont {Gastegger}, \citenamefont {Poltavsky}, \citenamefont {Sch{\"u}tt}, \citenamefont {Tkatchenko},\ and\ \citenamefont {M{\"u}ller}}]{Unke2021Machine}%
  \BibitemOpen
  \bibfield  {author} {\bibinfo {author} {\bibfnamefont {O.~T.}\ \bibnamefont {Unke}}, \bibinfo {author} {\bibfnamefont {S.}~\bibnamefont {Chmiela}}, \bibinfo {author} {\bibfnamefont {H.~E.}\ \bibnamefont {Sauceda}}, \bibinfo {author} {\bibfnamefont {M.}~\bibnamefont {Gastegger}}, \bibinfo {author} {\bibfnamefont {I.}~\bibnamefont {Poltavsky}}, \bibinfo {author} {\bibfnamefont {K.~T.}\ \bibnamefont {Sch{\"u}tt}}, \bibinfo {author} {\bibfnamefont {A.}~\bibnamefont {Tkatchenko}},\ and\ \bibinfo {author} {\bibfnamefont {K.-R.}\ \bibnamefont {M{\"u}ller}},\ }\bibfield  {title} {\bibinfo {title} {Machine learning force fields},\ }\href {https://doi.org/10.1021/acs.chemrev.0c01111} {\bibfield  {journal} {\bibinfo  {journal} {Chemical Reviews}\ }\textbf {\bibinfo {volume} {121}},\ \bibinfo {pages} {10142} (\bibinfo {year} {2021})}\BibitemShut {NoStop}%
\bibitem [{\citenamefont {Behler}\ and\ \citenamefont {Csányi}(2021)}]{Behler2021Machine}%
  \BibitemOpen
  \bibfield  {author} {\bibinfo {author} {\bibfnamefont {J.}~\bibnamefont {Behler}}\ and\ \bibinfo {author} {\bibfnamefont {G.}~\bibnamefont {Csányi}},\ }\bibfield  {title} {\bibinfo {title} {Machine learning potentials for extended systems: a perspective},\ }\href {https://doi.org/10.1140/epjb/s10051-021-00156-1} {\bibfield  {journal} {\bibinfo  {journal} {European Physical Journal B}\ }\textbf {\bibinfo {volume} {94}},\ \bibinfo {pages} {142} (\bibinfo {year} {2021})}\BibitemShut {NoStop}%
\bibitem [{\citenamefont {Friederich}\ \emph {et~al.}(2021)\citenamefont {Friederich}, \citenamefont {H{\"a}se}, \citenamefont {Proppe},\ and\ \citenamefont {{Aspuru-Guzik}}}]{Friederich2021Machine-Learned}%
  \BibitemOpen
  \bibfield  {author} {\bibinfo {author} {\bibfnamefont {P.}~\bibnamefont {Friederich}}, \bibinfo {author} {\bibfnamefont {F.}~\bibnamefont {H{\"a}se}}, \bibinfo {author} {\bibfnamefont {J.}~\bibnamefont {Proppe}},\ and\ \bibinfo {author} {\bibfnamefont {A.}~\bibnamefont {{Aspuru-Guzik}}},\ }\bibfield  {title} {\bibinfo {title} {Machine-learned potentials for next-generation matter simulations},\ }\href {https://doi.org/10.1038/s41563-020-0777-6} {\bibfield  {journal} {\bibinfo  {journal} {Nature Materials}\ }\textbf {\bibinfo {volume} {20}},\ \bibinfo {pages} {750} (\bibinfo {year} {2021})}\BibitemShut {NoStop}%
\bibitem [{\citenamefont {Behler}(2016)}]{Behler2016Perspective}%
  \BibitemOpen
  \bibfield  {author} {\bibinfo {author} {\bibfnamefont {J.}~\bibnamefont {Behler}},\ }\bibfield  {title} {\bibinfo {title} {Perspective: {{Machine}} learning potentials for atomistic simulations},\ }\href {https://doi.org/10.1063/1.4966192} {\bibfield  {journal} {\bibinfo  {journal} {Journal of Chemical Physics}\ }\textbf {\bibinfo {volume} {145}},\ \bibinfo {pages} {170901} (\bibinfo {year} {2016})}\BibitemShut {NoStop}%
\bibitem [{\citenamefont {Behler}\ and\ \citenamefont {Parrinello}(2007)}]{behler2007generalized}%
  \BibitemOpen
  \bibfield  {author} {\bibinfo {author} {\bibfnamefont {J.}~\bibnamefont {Behler}}\ and\ \bibinfo {author} {\bibfnamefont {M.}~\bibnamefont {Parrinello}},\ }\bibfield  {title} {\bibinfo {title} {{Generalized Neural-Network Representation of High-Dimensional Potential-Energy Surfaces}},\ }\href {https://doi.org/10.1103/PhysRevLett.98.146401} {\bibfield  {journal} {\bibinfo  {journal} {Physical Review Letters}\ }\textbf {\bibinfo {volume} {98}},\ \bibinfo {pages} {146401} (\bibinfo {year} {2007})}\BibitemShut {NoStop}%
\bibitem [{\citenamefont {Bart{\'o}k}\ \emph {et~al.}(2010)\citenamefont {Bart{\'o}k}, \citenamefont {Payne}, \citenamefont {Kondor},\ and\ \citenamefont {Cs{\'a}nyi}}]{bartok2010gaussian}%
  \BibitemOpen
  \bibfield  {author} {\bibinfo {author} {\bibfnamefont {A.~P.}\ \bibnamefont {Bart{\'o}k}}, \bibinfo {author} {\bibfnamefont {M.~C.}\ \bibnamefont {Payne}}, \bibinfo {author} {\bibfnamefont {R.}~\bibnamefont {Kondor}},\ and\ \bibinfo {author} {\bibfnamefont {G.}~\bibnamefont {Cs{\'a}nyi}},\ }\bibfield  {title} {\bibinfo {title} {{Gaussian approximation potentials: The accuracy of quantum mechanics, without the electrons}},\ }\href {https://doi.org/10.1103/PhysRevLett.104.136403} {\bibfield  {journal} {\bibinfo  {journal} {Physical Review Letters}\ }\textbf {\bibinfo {volume} {104}},\ \bibinfo {pages} {136403} (\bibinfo {year} {2010})}\BibitemShut {NoStop}%
\bibitem [{\citenamefont {Thompson}\ \emph {et~al.}(2015)\citenamefont {Thompson}, \citenamefont {Swiler}, \citenamefont {Trott}, \citenamefont {Foiles},\ and\ \citenamefont {Tucker}}]{Thompson2015jcp}%
  \BibitemOpen
  \bibfield  {author} {\bibinfo {author} {\bibfnamefont {A.}~\bibnamefont {Thompson}}, \bibinfo {author} {\bibfnamefont {L.}~\bibnamefont {Swiler}}, \bibinfo {author} {\bibfnamefont {C.}~\bibnamefont {Trott}}, \bibinfo {author} {\bibfnamefont {S.}~\bibnamefont {Foiles}},\ and\ \bibinfo {author} {\bibfnamefont {G.}~\bibnamefont {Tucker}},\ }\bibfield  {title} {\bibinfo {title} {{Spectral neighbor analysis method for automated generation of quantum-accurate interatomic potentials}},\ }\href {https://doi.org/https://doi.org/10.1016/j.jcp.2014.12.018} {\bibfield  {journal} {\bibinfo  {journal} {Journal of Computational Physics}\ }\textbf {\bibinfo {volume} {285}},\ \bibinfo {pages} {316} (\bibinfo {year} {2015})}\BibitemShut {NoStop}%
\bibitem [{\citenamefont {Shapeev}(2016)}]{shapeev2016moment}%
  \BibitemOpen
  \bibfield  {author} {\bibinfo {author} {\bibfnamefont {A.~V.}\ \bibnamefont {Shapeev}},\ }\bibfield  {title} {\bibinfo {title} {{Moment Tensor Potentials: A Class of Systematically Improvable Interatomic Potentials}},\ }\href {https://doi.org/10.1137/15M1054183} {\bibfield  {journal} {\bibinfo  {journal} {Multiscale Modeling \& Simulation}\ }\textbf {\bibinfo {volume} {14}},\ \bibinfo {pages} {1153} (\bibinfo {year} {2016})}\BibitemShut {NoStop}%
\bibitem [{\citenamefont {Wang}\ \emph {et~al.}(2018)\citenamefont {Wang}, \citenamefont {Zhang}, \citenamefont {Han},\ and\ \citenamefont {E}}]{Wang2018DeePMD-kit}%
  \BibitemOpen
  \bibfield  {author} {\bibinfo {author} {\bibfnamefont {H.}~\bibnamefont {Wang}}, \bibinfo {author} {\bibfnamefont {L.}~\bibnamefont {Zhang}}, \bibinfo {author} {\bibfnamefont {J.}~\bibnamefont {Han}},\ and\ \bibinfo {author} {\bibfnamefont {W.}~\bibnamefont {E}},\ }\bibfield  {title} {\bibinfo {title} {Deepmd-kit: A deep learning package for many-body potential energy representation and molecular dynamics},\ }\href {https://doi.org/https://doi.org/10.1016/j.cpc.2018.03.016} {\bibfield  {journal} {\bibinfo  {journal} {Computer Physics Communications}\ }\textbf {\bibinfo {volume} {228}},\ \bibinfo {pages} {178} (\bibinfo {year} {2018})}\BibitemShut {NoStop}%
\bibitem [{\citenamefont {Zhang}\ \emph {et~al.}(2018)\citenamefont {Zhang}, \citenamefont {Han}, \citenamefont {Wang}, \citenamefont {Car},\ and\ \citenamefont {E}}]{Zhang2018Deep}%
  \BibitemOpen
  \bibfield  {author} {\bibinfo {author} {\bibfnamefont {L.}~\bibnamefont {Zhang}}, \bibinfo {author} {\bibfnamefont {J.}~\bibnamefont {Han}}, \bibinfo {author} {\bibfnamefont {H.}~\bibnamefont {Wang}}, \bibinfo {author} {\bibfnamefont {R.}~\bibnamefont {Car}},\ and\ \bibinfo {author} {\bibfnamefont {W.}~\bibnamefont {E}},\ }\bibfield  {title} {\bibinfo {title} {Deep potential molecular dynamics: A scalable model with the accuracy of quantum mechanics},\ }\href {https://doi.org/10.1103/PhysRevLett.120.143001} {\bibfield  {journal} {\bibinfo  {journal} {Physical Review Letters}\ }\textbf {\bibinfo {volume} {120}},\ \bibinfo {pages} {143001} (\bibinfo {year} {2018})}\BibitemShut {NoStop}%
\bibitem [{\citenamefont {Drautz}(2019)}]{drautz2019atomic}%
  \BibitemOpen
  \bibfield  {author} {\bibinfo {author} {\bibfnamefont {R.}~\bibnamefont {Drautz}},\ }\bibfield  {title} {\bibinfo {title} {Atomic cluster expansion for accurate and transferable interatomic potentials},\ }\href {https://doi.org/10.1103/PhysRevB.99.014104} {\bibfield  {journal} {\bibinfo  {journal} {Physical Review B}\ }\textbf {\bibinfo {volume} {99}},\ \bibinfo {pages} {014104} (\bibinfo {year} {2019})}\BibitemShut {NoStop}%
\bibitem [{\citenamefont {Fan}\ \emph {et~al.}(2021)\citenamefont {Fan}, \citenamefont {Zeng}, \citenamefont {Zhang}, \citenamefont {Wang}, \citenamefont {Song}, \citenamefont {Dong}, \citenamefont {Chen},\ and\ \citenamefont {{Ala-Nissila}}}]{Fan2021Neuroevolution}%
  \BibitemOpen
  \bibfield  {author} {\bibinfo {author} {\bibfnamefont {Z.}~\bibnamefont {Fan}}, \bibinfo {author} {\bibfnamefont {Z.}~\bibnamefont {Zeng}}, \bibinfo {author} {\bibfnamefont {C.}~\bibnamefont {Zhang}}, \bibinfo {author} {\bibfnamefont {Y.}~\bibnamefont {Wang}}, \bibinfo {author} {\bibfnamefont {K.}~\bibnamefont {Song}}, \bibinfo {author} {\bibfnamefont {H.}~\bibnamefont {Dong}}, \bibinfo {author} {\bibfnamefont {Y.}~\bibnamefont {Chen}},\ and\ \bibinfo {author} {\bibfnamefont {T.}~\bibnamefont {{Ala-Nissila}}},\ }\bibfield  {title} {\bibinfo {title} {Neuroevolution machine learning potentials: {{Combining}} high accuracy and low cost in atomistic simulations and application to heat transport},\ }\href {https://doi.org/10.1103/PhysRevB.104.104309} {\bibfield  {journal} {\bibinfo  {journal} {Physical Review B}\ }\textbf {\bibinfo {volume} {104}},\ \bibinfo {pages} {104309} (\bibinfo {year} {2021})}\BibitemShut {NoStop}%
\bibitem [{\citenamefont {Fan}\ \emph {et~al.}(2022)\citenamefont {Fan}, \citenamefont {Wang}, \citenamefont {Ying}, \citenamefont {Song}, \citenamefont {Wang}, \citenamefont {Wang}, \citenamefont {Zeng}, \citenamefont {Xu}, \citenamefont {Lindgren}, \citenamefont {Rahm}, \citenamefont {Gabourie}, \citenamefont {Liu}, \citenamefont {Dong}, \citenamefont {Wu}, \citenamefont {Chen}, \citenamefont {Zhong}, \citenamefont {Sun}, \citenamefont {Erhart}, \citenamefont {Su},\ and\ \citenamefont {{Ala-Nissila}}}]{Fan2022GPUMD}%
  \BibitemOpen
  \bibfield  {author} {\bibinfo {author} {\bibfnamefont {Z.}~\bibnamefont {Fan}}, \bibinfo {author} {\bibfnamefont {Y.}~\bibnamefont {Wang}}, \bibinfo {author} {\bibfnamefont {P.}~\bibnamefont {Ying}}, \bibinfo {author} {\bibfnamefont {K.}~\bibnamefont {Song}}, \bibinfo {author} {\bibfnamefont {J.}~\bibnamefont {Wang}}, \bibinfo {author} {\bibfnamefont {Y.}~\bibnamefont {Wang}}, \bibinfo {author} {\bibfnamefont {Z.}~\bibnamefont {Zeng}}, \bibinfo {author} {\bibfnamefont {K.}~\bibnamefont {Xu}}, \bibinfo {author} {\bibfnamefont {E.}~\bibnamefont {Lindgren}}, \bibinfo {author} {\bibfnamefont {J.~M.}\ \bibnamefont {Rahm}}, \bibinfo {author} {\bibfnamefont {A.~J.}\ \bibnamefont {Gabourie}}, \bibinfo {author} {\bibfnamefont {J.}~\bibnamefont {Liu}}, \bibinfo {author} {\bibfnamefont {H.}~\bibnamefont {Dong}}, \bibinfo {author} {\bibfnamefont {J.}~\bibnamefont {Wu}}, \bibinfo {author} {\bibfnamefont {Y.}~\bibnamefont {Chen}}, \bibinfo {author} {\bibfnamefont {Z.}~\bibnamefont {Zhong}}, \bibinfo {author} {\bibfnamefont {J.}~\bibnamefont {Sun}}, \bibinfo {author} {\bibfnamefont {P.}~\bibnamefont {Erhart}}, \bibinfo {author} {\bibfnamefont {Y.}~\bibnamefont {Su}},\ and\ \bibinfo {author} {\bibfnamefont {T.}~\bibnamefont {{Ala-Nissila}}},\ }\bibfield  {title} {\bibinfo {title} {{{GPUMD}}: {{A}} package for constructing accurate machine-learned potentials and performing highly efficient atomistic simulations},\ }\href {https://doi.org/10.1063/5.0106617} {\bibfield  {journal} {\bibinfo  {journal} {Journal of Chemical Physics}\ }\textbf {\bibinfo {volume} {157}},\ \bibinfo {pages} {114801} (\bibinfo {year} {2022})}\BibitemShut {NoStop}%
\bibitem [{\citenamefont {Fan}\ \emph {et~al.}(2017)\citenamefont {Fan}, \citenamefont {Chen}, \citenamefont {Vierimaa},\ and\ \citenamefont {Harju}}]{fan2017efficient}%
  \BibitemOpen
  \bibfield  {author} {\bibinfo {author} {\bibfnamefont {Z.}~\bibnamefont {Fan}}, \bibinfo {author} {\bibfnamefont {W.}~\bibnamefont {Chen}}, \bibinfo {author} {\bibfnamefont {V.}~\bibnamefont {Vierimaa}},\ and\ \bibinfo {author} {\bibfnamefont {A.}~\bibnamefont {Harju}},\ }\bibfield  {title} {\bibinfo {title} {{Efficient molecular dynamics simulations with many-body potentials on graphics processing units}},\ }\href {https://doi.org/https://doi.org/10.1016/j.cpc.2017.05.003} {\bibfield  {journal} {\bibinfo  {journal} {Computer Physics Communications}\ }\textbf {\bibinfo {volume} {218}},\ \bibinfo {pages} {10} (\bibinfo {year} {2017})}\BibitemShut {NoStop}%
\bibitem [{\citenamefont {Dong}\ \emph {et~al.}(2024)\citenamefont {Dong}, \citenamefont {Shi}, \citenamefont {Ying}, \citenamefont {Xu}, \citenamefont {Liang}, \citenamefont {Wang}, \citenamefont {Zeng}, \citenamefont {Wu}, \citenamefont {Zhou}, \citenamefont {Xiong}, \citenamefont {Chen},\ and\ \citenamefont {Fan}}]{dong2024molecular}%
  \BibitemOpen
  \bibfield  {author} {\bibinfo {author} {\bibfnamefont {H.}~\bibnamefont {Dong}}, \bibinfo {author} {\bibfnamefont {Y.}~\bibnamefont {Shi}}, \bibinfo {author} {\bibfnamefont {P.}~\bibnamefont {Ying}}, \bibinfo {author} {\bibfnamefont {K.}~\bibnamefont {Xu}}, \bibinfo {author} {\bibfnamefont {T.}~\bibnamefont {Liang}}, \bibinfo {author} {\bibfnamefont {Y.}~\bibnamefont {Wang}}, \bibinfo {author} {\bibfnamefont {Z.}~\bibnamefont {Zeng}}, \bibinfo {author} {\bibfnamefont {X.}~\bibnamefont {Wu}}, \bibinfo {author} {\bibfnamefont {W.}~\bibnamefont {Zhou}}, \bibinfo {author} {\bibfnamefont {S.}~\bibnamefont {Xiong}}, \bibinfo {author} {\bibfnamefont {S.}~\bibnamefont {Chen}},\ and\ \bibinfo {author} {\bibfnamefont {Z.}~\bibnamefont {Fan}},\ }\bibfield  {title} {\bibinfo {title} {Molecular dynamics simulations of heat transport using machine-learned potentials: A mini-review and tutorial on gpumd with neuroevolution potentials},\ }\href {https://doi.org/10.1063/5.0200833} {\bibfield  {journal} {\bibinfo  {journal} {Journal of Applied Physics}\ }\textbf {\bibinfo {volume} {135}},\ \bibinfo {pages} {161101} (\bibinfo {year} {2024})}\BibitemShut {NoStop}%
\bibitem [{\citenamefont {Ying}\ \emph {et~al.}(2025)\citenamefont {Ying}, \citenamefont {Qian}, \citenamefont {Zhao}, \citenamefont {Wang}, \citenamefont {Xu}, \citenamefont {Ding}, \citenamefont {Chen},\ and\ \citenamefont {Fan}}]{ying2025advances}%
  \BibitemOpen
  \bibfield  {author} {\bibinfo {author} {\bibfnamefont {P.}~\bibnamefont {Ying}}, \bibinfo {author} {\bibfnamefont {C.}~\bibnamefont {Qian}}, \bibinfo {author} {\bibfnamefont {R.}~\bibnamefont {Zhao}}, \bibinfo {author} {\bibfnamefont {Y.}~\bibnamefont {Wang}}, \bibinfo {author} {\bibfnamefont {K.}~\bibnamefont {Xu}}, \bibinfo {author} {\bibfnamefont {F.}~\bibnamefont {Ding}}, \bibinfo {author} {\bibfnamefont {S.}~\bibnamefont {Chen}},\ and\ \bibinfo {author} {\bibfnamefont {Z.}~\bibnamefont {Fan}},\ }\bibfield  {title} {\bibinfo {title} {Advances in modeling complex materials: The rise of neuroevolution potentials},\ }\href {https://doi.org/10.1063/5.0259061} {\bibfield  {journal} {\bibinfo  {journal} {Chemical Physics Reviews}\ }\textbf {\bibinfo {volume} {6}},\ \bibinfo {pages} {011310} (\bibinfo {year} {2025})}\BibitemShut {NoStop}%
\bibitem [{\citenamefont {Wang}\ \emph {et~al.}(2024)\citenamefont {Wang}, \citenamefont {Fan}, \citenamefont {Qian}, \citenamefont {Caro},\ and\ \citenamefont {{Ala-Nissila}}}]{Wang2024Density}%
  \BibitemOpen
  \bibfield  {author} {\bibinfo {author} {\bibfnamefont {Y.}~\bibnamefont {Wang}}, \bibinfo {author} {\bibfnamefont {Z.}~\bibnamefont {Fan}}, \bibinfo {author} {\bibfnamefont {P.}~\bibnamefont {Qian}}, \bibinfo {author} {\bibfnamefont {M.~A.}\ \bibnamefont {Caro}},\ and\ \bibinfo {author} {\bibfnamefont {T.}~\bibnamefont {{Ala-Nissila}}},\ }\href {https://doi.org/10.48550/arXiv.2408.12390} {\bibinfo {title} {Density dependence of thermal conductivity in nanoporous and amorphous carbon with machine-learned molecular dynamics}} (\bibinfo {year} {2024}),\ \Eprint {https://arxiv.org/abs/2408.12390} {arXiv:2408.12390 [cond-mat]} \BibitemShut {NoStop}%
\bibitem [{\citenamefont {Xu}\ \emph {et~al.}(2024)\citenamefont {Xu}, \citenamefont {Liang}, \citenamefont {Xu}, \citenamefont {Ying}, \citenamefont {Chen}, \citenamefont {Wei}, \citenamefont {Xu},\ and\ \citenamefont {Fan}}]{Xu2024NEPMBpol}%
  \BibitemOpen
  \bibfield  {author} {\bibinfo {author} {\bibfnamefont {K.}~\bibnamefont {Xu}}, \bibinfo {author} {\bibfnamefont {T.}~\bibnamefont {Liang}}, \bibinfo {author} {\bibfnamefont {N.}~\bibnamefont {Xu}}, \bibinfo {author} {\bibfnamefont {P.}~\bibnamefont {Ying}}, \bibinfo {author} {\bibfnamefont {S.}~\bibnamefont {Chen}}, \bibinfo {author} {\bibfnamefont {N.}~\bibnamefont {Wei}}, \bibinfo {author} {\bibfnamefont {J.}~\bibnamefont {Xu}},\ and\ \bibinfo {author} {\bibfnamefont {Z.}~\bibnamefont {Fan}},\ }\href {https://doi.org/10.48550/arXiv.2411.09631} {\bibinfo {title} {{{NEP-MB-pol}}: {{A}} unified machine-learned framework for fast and accurate prediction of water's thermodynamic and transport properties}} (\bibinfo {year} {2024}),\ \Eprint {https://arxiv.org/abs/2411.09631} {arXiv:2411.09631 [physics]} \BibitemShut {NoStop}%
\bibitem [{\citenamefont {Chen}\ \emph {et~al.}(2024)\citenamefont {Chen}, \citenamefont {Jin}, \citenamefont {Zhao},\ and\ \citenamefont {Li}}]{Chen2024Intricate}%
  \BibitemOpen
  \bibfield  {author} {\bibinfo {author} {\bibfnamefont {S.}~\bibnamefont {Chen}}, \bibinfo {author} {\bibfnamefont {X.}~\bibnamefont {Jin}}, \bibinfo {author} {\bibfnamefont {W.}~\bibnamefont {Zhao}},\ and\ \bibinfo {author} {\bibfnamefont {T.}~\bibnamefont {Li}},\ }\bibfield  {title} {\bibinfo {title} {Intricate short-range order in {{GeSn}} alloys revealed by atomistic simulations with highly accurate and efficient machine-learning potentials},\ }\href {https://doi.org/10.1103/PhysRevMaterials.8.043805} {\bibfield  {journal} {\bibinfo  {journal} {Physical Review Materials}\ }\textbf {\bibinfo {volume} {8}},\ \bibinfo {pages} {043805} (\bibinfo {year} {2024})}\BibitemShut {NoStop}%
\bibitem [{\citenamefont {Fransson}\ \emph {et~al.}(2023)\citenamefont {Fransson}, \citenamefont {Wiktor},\ and\ \citenamefont {Erhart}}]{Fransson2023Phase}%
  \BibitemOpen
  \bibfield  {author} {\bibinfo {author} {\bibfnamefont {E.}~\bibnamefont {Fransson}}, \bibinfo {author} {\bibfnamefont {J.}~\bibnamefont {Wiktor}},\ and\ \bibinfo {author} {\bibfnamefont {P.}~\bibnamefont {Erhart}},\ }\bibfield  {title} {\bibinfo {title} {Phase transitions in inorganic halide perovskites from machine-learned potentials},\ }\href {https://doi.org/10.1021/acs.jpcc.3c01542} {\bibfield  {journal} {\bibinfo  {journal} {The Journal of Physical Chemistry C}\ }\textbf {\bibinfo {volume} {127}},\ \bibinfo {pages} {13773} (\bibinfo {year} {2023})}\BibitemShut {NoStop}%
\bibitem [{\citenamefont {Wiktor}\ \emph {et~al.}(2023)\citenamefont {Wiktor}, \citenamefont {Fransson}, \citenamefont {Kubicki},\ and\ \citenamefont {Erhart}}]{Wiktor2023Quantifying}%
  \BibitemOpen
  \bibfield  {author} {\bibinfo {author} {\bibfnamefont {J.}~\bibnamefont {Wiktor}}, \bibinfo {author} {\bibfnamefont {E.}~\bibnamefont {Fransson}}, \bibinfo {author} {\bibfnamefont {D.}~\bibnamefont {Kubicki}},\ and\ \bibinfo {author} {\bibfnamefont {P.}~\bibnamefont {Erhart}},\ }\bibfield  {title} {\bibinfo {title} {Quantifying dynamic tilting in halide perovskites: {{Chemical}} trends and local correlations},\ }\href {https://doi.org/10.1021/acs.chemmater.3c00933} {\bibfield  {journal} {\bibinfo  {journal} {Chemistry of Materials}\ }\textbf {\bibinfo {volume} {35}},\ \bibinfo {pages} {6737} (\bibinfo {year} {2023})}\BibitemShut {NoStop}%
\bibitem [{\citenamefont {Qian}\ \emph {et~al.}(2024)\citenamefont {Qian}, \citenamefont {Hedman}, \citenamefont {Li}, \citenamefont {Kim},\ and\ \citenamefont {Ding}}]{Qian2024Reconstruction}%
  \BibitemOpen
  \bibfield  {author} {\bibinfo {author} {\bibfnamefont {C.}~\bibnamefont {Qian}}, \bibinfo {author} {\bibfnamefont {D.}~\bibnamefont {Hedman}}, \bibinfo {author} {\bibfnamefont {P.}~\bibnamefont {Li}}, \bibinfo {author} {\bibfnamefont {S.~Y.}\ \bibnamefont {Kim}},\ and\ \bibinfo {author} {\bibfnamefont {F.}~\bibnamefont {Ding}},\ }\bibfield  {title} {\bibinfo {title} {The reconstruction of {{Pt}}(001) surface and the shell-like reconstruction of the vicinal {{Pt}}(001) surfaces revealed by neural network potential},\ }\href {https://doi.org/10.1002/smll.202404274} {\bibfield  {journal} {\bibinfo  {journal} {Small}\ }\textbf {\bibinfo {volume} {20}},\ \bibinfo {pages} {2404274} (\bibinfo {year} {2024})}\BibitemShut {NoStop}%
\bibitem [{\citenamefont {Liu}\ \emph {et~al.}(2023)\citenamefont {Liu}, \citenamefont {Byggm{\"a}star}, \citenamefont {Fan}, \citenamefont {Qian},\ and\ \citenamefont {Su}}]{Liu2023Largescale}%
  \BibitemOpen
  \bibfield  {author} {\bibinfo {author} {\bibfnamefont {J.}~\bibnamefont {Liu}}, \bibinfo {author} {\bibfnamefont {J.}~\bibnamefont {Byggm{\"a}star}}, \bibinfo {author} {\bibfnamefont {Z.}~\bibnamefont {Fan}}, \bibinfo {author} {\bibfnamefont {P.}~\bibnamefont {Qian}},\ and\ \bibinfo {author} {\bibfnamefont {Y.}~\bibnamefont {Su}},\ }\bibfield  {title} {\bibinfo {title} {Large-scale machine-learning molecular dynamics simulation of primary radiation damage in tungsten},\ }\href {https://doi.org/10.1103/PhysRevB.108.054312} {\bibfield  {journal} {\bibinfo  {journal} {Physical Review B}\ }\textbf {\bibinfo {volume} {108}},\ \bibinfo {pages} {054312} (\bibinfo {year} {2023})}\BibitemShut {NoStop}%
\bibitem [{\citenamefont {Ying}\ \emph {et~al.}(2023)\citenamefont {Ying}, \citenamefont {Dong}, \citenamefont {Liang}, \citenamefont {Fan}, \citenamefont {Zhong},\ and\ \citenamefont {Zhang}}]{Ying2023Atomistic}%
  \BibitemOpen
  \bibfield  {author} {\bibinfo {author} {\bibfnamefont {P.}~\bibnamefont {Ying}}, \bibinfo {author} {\bibfnamefont {H.}~\bibnamefont {Dong}}, \bibinfo {author} {\bibfnamefont {T.}~\bibnamefont {Liang}}, \bibinfo {author} {\bibfnamefont {Z.}~\bibnamefont {Fan}}, \bibinfo {author} {\bibfnamefont {Z.}~\bibnamefont {Zhong}},\ and\ \bibinfo {author} {\bibfnamefont {J.}~\bibnamefont {Zhang}},\ }\bibfield  {title} {\bibinfo {title} {Atomistic insights into the mechanical anisotropy and fragility of monolayer fullerene networks using quantum mechanical calculations and machine-learning molecular dynamics simulations},\ }\href {https://doi.org/10.1016/j.eml.2022.101929} {\bibfield  {journal} {\bibinfo  {journal} {Extreme Mechanics Letters}\ }\textbf {\bibinfo {volume} {58}},\ \bibinfo {pages} {101929} (\bibinfo {year} {2023})}\BibitemShut {NoStop}%
\bibitem [{\citenamefont {Yu}\ \emph {et~al.}(2024)\citenamefont {Yu}, \citenamefont {Zhao}, \citenamefont {Guo},\ and\ \citenamefont {Zhang}}]{Yu2024Fracture}%
  \BibitemOpen
  \bibfield  {author} {\bibinfo {author} {\bibfnamefont {M.}~\bibnamefont {Yu}}, \bibinfo {author} {\bibfnamefont {Z.}~\bibnamefont {Zhao}}, \bibinfo {author} {\bibfnamefont {W.}~\bibnamefont {Guo}},\ and\ \bibinfo {author} {\bibfnamefont {Z.}~\bibnamefont {Zhang}},\ }\bibfield  {title} {\bibinfo {title} {Fracture toughness of two-dimensional materials dominated by edge energy anisotropy},\ }\href {https://doi.org/10.1016/j.jmps.2024.105579} {\bibfield  {journal} {\bibinfo  {journal} {Journal of the Mechanics and Physics of Solids}\ }\textbf {\bibinfo {volume} {186}},\ \bibinfo {pages} {105579} (\bibinfo {year} {2024})}\BibitemShut {NoStop}%
\bibitem [{\citenamefont {Song}\ \emph {et~al.}(2024)\citenamefont {Song}, \citenamefont {Zhao}, \citenamefont {Liu}, \citenamefont {Wang}, \citenamefont {Lindgren}, \citenamefont {Wang}, \citenamefont {Chen}, \citenamefont {Xu}, \citenamefont {Liang}, \citenamefont {Ying}, \citenamefont {Xu}, \citenamefont {Zhao}, \citenamefont {Shi}, \citenamefont {Wang}, \citenamefont {Lyu}, \citenamefont {Zeng}, \citenamefont {Liang}, \citenamefont {Dong}, \citenamefont {Sun}, \citenamefont {Chen}, \citenamefont {Zhang}, \citenamefont {Guo}, \citenamefont {Qian}, \citenamefont {Sun}, \citenamefont {Erhart}, \citenamefont {Ala-Nissila}, \citenamefont {Su},\ and\ \citenamefont {Fan}}]{Song2024General}%
  \BibitemOpen
  \bibfield  {author} {\bibinfo {author} {\bibfnamefont {K.}~\bibnamefont {Song}}, \bibinfo {author} {\bibfnamefont {R.}~\bibnamefont {Zhao}}, \bibinfo {author} {\bibfnamefont {J.}~\bibnamefont {Liu}}, \bibinfo {author} {\bibfnamefont {Y.}~\bibnamefont {Wang}}, \bibinfo {author} {\bibfnamefont {E.}~\bibnamefont {Lindgren}}, \bibinfo {author} {\bibfnamefont {Y.}~\bibnamefont {Wang}}, \bibinfo {author} {\bibfnamefont {S.}~\bibnamefont {Chen}}, \bibinfo {author} {\bibfnamefont {K.}~\bibnamefont {Xu}}, \bibinfo {author} {\bibfnamefont {T.}~\bibnamefont {Liang}}, \bibinfo {author} {\bibfnamefont {P.}~\bibnamefont {Ying}}, \bibinfo {author} {\bibfnamefont {N.}~\bibnamefont {Xu}}, \bibinfo {author} {\bibfnamefont {Z.}~\bibnamefont {Zhao}}, \bibinfo {author} {\bibfnamefont {J.}~\bibnamefont {Shi}}, \bibinfo {author} {\bibfnamefont {J.}~\bibnamefont {Wang}}, \bibinfo {author} {\bibfnamefont {S.}~\bibnamefont {Lyu}}, \bibinfo {author} {\bibfnamefont {Z.}~\bibnamefont {Zeng}}, \bibinfo {author} {\bibfnamefont {S.}~\bibnamefont {Liang}}, \bibinfo {author} {\bibfnamefont {H.}~\bibnamefont {Dong}}, \bibinfo {author} {\bibfnamefont {L.}~\bibnamefont {Sun}}, \bibinfo {author} {\bibfnamefont {Y.}~\bibnamefont {Chen}}, \bibinfo {author} {\bibfnamefont {Z.}~\bibnamefont {Zhang}}, \bibinfo {author} {\bibfnamefont {W.}~\bibnamefont {Guo}}, \bibinfo {author} {\bibfnamefont {P.}~\bibnamefont {Qian}}, \bibinfo {author} {\bibfnamefont {J.}~\bibnamefont {Sun}}, \bibinfo {author} {\bibfnamefont {P.}~\bibnamefont {Erhart}}, \bibinfo {author} {\bibfnamefont {T.}~\bibnamefont {Ala-Nissila}}, \bibinfo {author} {\bibfnamefont {Y.}~\bibnamefont {Su}},\ and\ \bibinfo {author} {\bibfnamefont {Z.}~\bibnamefont {Fan}},\ }\bibfield  {title} {\bibinfo {title} {{General-purpose machine-learned potential for 16 elemental metals and their alloys}},\ }\href {https://doi.org/10.1038/s41467-024-54554-x} {\bibfield  {journal} {\bibinfo  {journal} {Nature Communications}\ }\textbf {\bibinfo {volume} {15}},\ \bibinfo {pages} {10208} (\bibinfo {year} {2024})}\BibitemShut {NoStop}%
\bibitem [{\citenamefont {metatensor}(2025)}]{Metatensor2025Metatrain}%
  \BibitemOpen
  \bibfield  {author} {\bibinfo {author} {\bibnamefont {metatensor}},\ }\href@noop {} {\bibinfo {title} {metatrain}},\ \bibinfo {howpublished} {\url{https://github.com/metatensor/metatrain}} (\bibinfo {year} {2025})\BibitemShut {NoStop}%
\bibitem [{\citenamefont {Sch{\"a}fer}\ \emph {et~al.}(2025)\citenamefont {Sch{\"a}fer}, \citenamefont {Segreto}, \citenamefont {Zills}, \citenamefont {Holm},\ and\ \citenamefont {K{\"a}stner}}]{Schafer2025Apax}%
  \BibitemOpen
  \bibfield  {author} {\bibinfo {author} {\bibfnamefont {M.~R.}\ \bibnamefont {Sch{\"a}fer}}, \bibinfo {author} {\bibfnamefont {N.}~\bibnamefont {Segreto}}, \bibinfo {author} {\bibfnamefont {F.}~\bibnamefont {Zills}}, \bibinfo {author} {\bibfnamefont {C.}~\bibnamefont {Holm}},\ and\ \bibinfo {author} {\bibfnamefont {J.}~\bibnamefont {K{\"a}stner}},\ }\href {https://doi.org/10.48550/arXiv.2505.22168} {\bibinfo {title} {Apax: {{A}} flexible and performant framework for the development of machine-learned interatomic potentials}} (\bibinfo {year} {2025}),\ \Eprint {https://arxiv.org/abs/2505.22168} {arXiv:2505.22168 [physics]} \BibitemShut {NoStop}%
\bibitem [{\citenamefont {Benner}(2025)}]{Philipp2025Equitrain}%
  \BibitemOpen
  \bibfield  {author} {\bibinfo {author} {\bibfnamefont {P.}~\bibnamefont {Benner}},\ }\href@noop {} {\bibinfo {title} {equitrain}},\ \bibinfo {howpublished} {\url{https://github.com/BAMeScience/equitrain}} (\bibinfo {year} {2025})\BibitemShut {NoStop}%
\bibitem [{\citenamefont {Brunken}\ \emph {et~al.}(2025)\citenamefont {Brunken}, \citenamefont {Peltre}, \citenamefont {Chomet}, \citenamefont {Walewski}, \citenamefont {McAuliffe}, \citenamefont {Heyraud}, \citenamefont {Attias}, \citenamefont {Maarand}, \citenamefont {Khanfir}, \citenamefont {Toledo}, \citenamefont {Falcioni}, \citenamefont {Bluntzer}, \citenamefont {{Acosta-Guti{\'e}rrez}},\ and\ \citenamefont {Tilly}}]{Brunken2025MachineLearning}%
  \BibitemOpen
  \bibfield  {author} {\bibinfo {author} {\bibfnamefont {C.}~\bibnamefont {Brunken}}, \bibinfo {author} {\bibfnamefont {O.}~\bibnamefont {Peltre}}, \bibinfo {author} {\bibfnamefont {H.}~\bibnamefont {Chomet}}, \bibinfo {author} {\bibfnamefont {L.}~\bibnamefont {Walewski}}, \bibinfo {author} {\bibfnamefont {M.}~\bibnamefont {McAuliffe}}, \bibinfo {author} {\bibfnamefont {V.}~\bibnamefont {Heyraud}}, \bibinfo {author} {\bibfnamefont {S.}~\bibnamefont {Attias}}, \bibinfo {author} {\bibfnamefont {M.}~\bibnamefont {Maarand}}, \bibinfo {author} {\bibfnamefont {Y.}~\bibnamefont {Khanfir}}, \bibinfo {author} {\bibfnamefont {E.}~\bibnamefont {Toledo}}, \bibinfo {author} {\bibfnamefont {F.}~\bibnamefont {Falcioni}}, \bibinfo {author} {\bibfnamefont {M.}~\bibnamefont {Bluntzer}}, \bibinfo {author} {\bibfnamefont {S.}~\bibnamefont {{Acosta-Guti{\'e}rrez}}},\ and\ \bibinfo {author} {\bibfnamefont {J.}~\bibnamefont {Tilly}},\ }\href {https://doi.org/10.48550/arXiv.2505.22397} {\bibinfo {title} {Machine learning interatomic potentials: {{Library}} for efficient training, model development and simulation of molecular systems}} (\bibinfo {year} {2025}),\ \Eprint {https://arxiv.org/abs/2505.22397} {arXiv:2505.22397 [physics]} \BibitemShut {NoStop}%
\bibitem [{\citenamefont {Liu}\ \emph {et~al.}(2024)\citenamefont {Liu}, \citenamefont {Morrow}, \citenamefont {Ertural}, \citenamefont {Fragapane}, \citenamefont {Gardner}, \citenamefont {Naik}, \citenamefont {Zhou}, \citenamefont {George},\ and\ \citenamefont {Deringer}}]{Liu2024Automated}%
  \BibitemOpen
  \bibfield  {author} {\bibinfo {author} {\bibfnamefont {Y.}~\bibnamefont {Liu}}, \bibinfo {author} {\bibfnamefont {J.~D.}\ \bibnamefont {Morrow}}, \bibinfo {author} {\bibfnamefont {C.}~\bibnamefont {Ertural}}, \bibinfo {author} {\bibfnamefont {N.~L.}\ \bibnamefont {Fragapane}}, \bibinfo {author} {\bibfnamefont {J.~L.~A.}\ \bibnamefont {Gardner}}, \bibinfo {author} {\bibfnamefont {A.~A.}\ \bibnamefont {Naik}}, \bibinfo {author} {\bibfnamefont {Y.}~\bibnamefont {Zhou}}, \bibinfo {author} {\bibfnamefont {J.}~\bibnamefont {George}},\ and\ \bibinfo {author} {\bibfnamefont {V.~L.}\ \bibnamefont {Deringer}},\ }\href {https://doi.org/10.48550/arXiv.2412.16736} {\bibinfo {title} {An automated framework for exploring and learning potential-energy surfaces}} (\bibinfo {year} {2024}),\ \Eprint {https://arxiv.org/abs/2412.16736} {arXiv:2412.16736 [physics]} \BibitemShut {NoStop}%
\bibitem [{\citenamefont {Zhang}\ \emph {et~al.}(2020)\citenamefont {Zhang}, \citenamefont {Wang}, \citenamefont {Chen}, \citenamefont {Zeng}, \citenamefont {Zhang}, \citenamefont {Wang},\ and\ \citenamefont {E}}]{Zhang2020DPGEN}%
  \BibitemOpen
  \bibfield  {author} {\bibinfo {author} {\bibfnamefont {Y.}~\bibnamefont {Zhang}}, \bibinfo {author} {\bibfnamefont {H.}~\bibnamefont {Wang}}, \bibinfo {author} {\bibfnamefont {W.}~\bibnamefont {Chen}}, \bibinfo {author} {\bibfnamefont {J.}~\bibnamefont {Zeng}}, \bibinfo {author} {\bibfnamefont {L.}~\bibnamefont {Zhang}}, \bibinfo {author} {\bibfnamefont {H.}~\bibnamefont {Wang}},\ and\ \bibinfo {author} {\bibfnamefont {W.}~\bibnamefont {E}},\ }\bibfield  {title} {\bibinfo {title} {{{DP-GEN}}: {{A}} concurrent learning platform for the generation of reliable deep learning based potential energy models},\ }\href {https://doi.org/10.1016/j.cpc.2020.107206} {\bibfield  {journal} {\bibinfo  {journal} {Computer Physics Communications}\ }\textbf {\bibinfo {volume} {253}},\ \bibinfo {pages} {107206} (\bibinfo {year} {2020})}\BibitemShut {NoStop}%
\bibitem [{\citenamefont {Galib}\ \emph {et~al.}(2025)\citenamefont {Galib}, \citenamefont {Isiet},\ and\ \citenamefont {Ponga}}]{Galib2025AtomProNet}%
  \BibitemOpen
  \bibfield  {author} {\bibinfo {author} {\bibfnamefont {M.}~\bibnamefont {Galib}}, \bibinfo {author} {\bibfnamefont {M.}~\bibnamefont {Isiet}},\ and\ \bibinfo {author} {\bibfnamefont {M.}~\bibnamefont {Ponga}},\ }\href {https://doi.org/10.48550/arXiv.2501.14039} {\bibinfo {title} {{{AtomProNet}}: {{Data}} flow to and from machine learning interatomic potentials in materials science}} (\bibinfo {year} {2025}),\ \Eprint {https://arxiv.org/abs/2501.14039} {arXiv:2501.14039 [cond-mat]} \BibitemShut {NoStop}%
\bibitem [{\citenamefont {Batatia}\ \emph {et~al.}(2022)\citenamefont {Batatia}, \citenamefont {Kovacs}, \citenamefont {Simm}, \citenamefont {Ortner},\ and\ \citenamefont {Csanyi}}]{Batatia2022Mace}%
  \BibitemOpen
  \bibfield  {author} {\bibinfo {author} {\bibfnamefont {I.}~\bibnamefont {Batatia}}, \bibinfo {author} {\bibfnamefont {D.~P.}\ \bibnamefont {Kovacs}}, \bibinfo {author} {\bibfnamefont {G.}~\bibnamefont {Simm}}, \bibinfo {author} {\bibfnamefont {C.}~\bibnamefont {Ortner}},\ and\ \bibinfo {author} {\bibfnamefont {G.}~\bibnamefont {Csanyi}},\ }\bibfield  {title} {\bibinfo {title} {{{MACE}}: {{Higher}} order equivariant message passing neural networks for fast and accurate force fields},\ }in\ \href@noop {} {\emph {\bibinfo {booktitle} {Advances in Neural Information Processing Systems}}},\ Vol.~\bibinfo {volume} {35},\ \bibinfo {editor} {edited by\ \bibinfo {editor} {\bibfnamefont {S.}~\bibnamefont {Koyejo}}, \bibinfo {editor} {\bibfnamefont {S.}~\bibnamefont {Mohamed}}, \bibinfo {editor} {\bibfnamefont {A.}~\bibnamefont {Agarwal}}, \bibinfo {editor} {\bibfnamefont {D.}~\bibnamefont {Belgrave}}, \bibinfo {editor} {\bibfnamefont {K.}~\bibnamefont {Cho}},\ and\ \bibinfo {editor} {\bibfnamefont {A.}~\bibnamefont {Oh}}}\ (\bibinfo  {publisher} {Curran Associates, Inc.},\ \bibinfo {year} {2022})\ pp.\ \bibinfo {pages} {11423--11436}\BibitemShut {NoStop}%
\bibitem [{\citenamefont {Lindgren}\ \emph {et~al.}(2024)\citenamefont {Lindgren}, \citenamefont {Rahm}, \citenamefont {Fransson}, \citenamefont {Eriksson}, \citenamefont {Österbacka}, \citenamefont {Fan},\ and\ \citenamefont {Erhart}}]{lindgren2024calorine}%
  \BibitemOpen
  \bibfield  {author} {\bibinfo {author} {\bibfnamefont {E.}~\bibnamefont {Lindgren}}, \bibinfo {author} {\bibfnamefont {M.}~\bibnamefont {Rahm}}, \bibinfo {author} {\bibfnamefont {E.}~\bibnamefont {Fransson}}, \bibinfo {author} {\bibfnamefont {F.}~\bibnamefont {Eriksson}}, \bibinfo {author} {\bibfnamefont {N.}~\bibnamefont {Österbacka}}, \bibinfo {author} {\bibfnamefont {Z.}~\bibnamefont {Fan}},\ and\ \bibinfo {author} {\bibfnamefont {P.}~\bibnamefont {Erhart}},\ }\bibfield  {title} {\bibinfo {title} {{calorine: A Python package for constructing and sampling neuroevolution potential models}},\ }\href {https://doi.org/10.21105/joss.06264} {\bibfield  {journal} {\bibinfo  {journal} {Journal of Open Source Software}\ }\textbf {\bibinfo {volume} {9}},\ \bibinfo {pages} {6264} (\bibinfo {year} {2024})}\BibitemShut {NoStop}%
\bibitem [{\citenamefont {Kresse}\ and\ \citenamefont {Hafner}(1994)}]{Kresse1994AbInitio}%
  \BibitemOpen
  \bibfield  {author} {\bibinfo {author} {\bibfnamefont {G.}~\bibnamefont {Kresse}}\ and\ \bibinfo {author} {\bibfnamefont {J.}~\bibnamefont {Hafner}},\ }\bibfield  {title} {\bibinfo {title} {{\emph{Ab Initio}} molecular-dynamics simulation of the liquid-metal--amorphous-semiconductor transition in germanium},\ }\href {https://doi.org/10.1103/PhysRevB.49.14251} {\bibfield  {journal} {\bibinfo  {journal} {Physical Review B}\ }\textbf {\bibinfo {volume} {49}},\ \bibinfo {pages} {14251} (\bibinfo {year} {1994})}\BibitemShut {NoStop}%
\bibitem [{\citenamefont {Kresse}\ and\ \citenamefont {Furthm{\"u}ller}(1996)}]{Kresse1996Efficient}%
  \BibitemOpen
  \bibfield  {author} {\bibinfo {author} {\bibfnamefont {G.}~\bibnamefont {Kresse}}\ and\ \bibinfo {author} {\bibfnamefont {J.}~\bibnamefont {Furthm{\"u}ller}},\ }\bibfield  {title} {\bibinfo {title} {Efficient iterative schemes for {\emph{ab initio}} total-energy calculations using a plane-wave basis set},\ }\href {https://doi.org/10.1103/PhysRevB.54.11169} {\bibfield  {journal} {\bibinfo  {journal} {Physical Review B}\ }\textbf {\bibinfo {volume} {54}},\ \bibinfo {pages} {11169} (\bibinfo {year} {1996})}\BibitemShut {NoStop}%
\bibitem [{\citenamefont {Schaul}\ \emph {et~al.}(2011)\citenamefont {Schaul}, \citenamefont {Glasmachers},\ and\ \citenamefont {Schmidhuber}}]{schaul2011high}%
  \BibitemOpen
  \bibfield  {author} {\bibinfo {author} {\bibfnamefont {T.}~\bibnamefont {Schaul}}, \bibinfo {author} {\bibfnamefont {T.}~\bibnamefont {Glasmachers}},\ and\ \bibinfo {author} {\bibfnamefont {J.}~\bibnamefont {Schmidhuber}},\ }\bibfield  {title} {\bibinfo {title} {{High Dimensions and Heavy Tails for Natural Evolution Strategies}},\ }in\ \href {https://doi.org/10.1145/2001576.2001692} {\emph {\bibinfo {booktitle} {Proceedings of the 13th Annual Conference on Genetic and Evolutionary Computation}}},\ \bibinfo {series and number} {GECCO '11}\ (\bibinfo  {publisher} {Association for Computing Machinery},\ \bibinfo {address} {New York, NY, USA},\ \bibinfo {year} {2011})\ pp.\ \bibinfo {pages} {845--852}\BibitemShut {NoStop}%
\bibitem [{\citenamefont {Liang}\ \emph {et~al.}(2025)\citenamefont {Liang}, \citenamefont {Xu}, \citenamefont {Lindgren}, \citenamefont {Chen}, \citenamefont {Zhao}, \citenamefont {Liu}, \citenamefont {Tang}, \citenamefont {Zhang}, \citenamefont {Wang}, \citenamefont {Song}, \citenamefont {Ying}, \citenamefont {Dong}, \citenamefont {Chen}, \citenamefont {Erhart}, \citenamefont {Fan}, \citenamefont {Ala-Nissila},\ and\ \citenamefont {Xu}}]{liang2025nep89}%
  \BibitemOpen
  \bibfield  {author} {\bibinfo {author} {\bibfnamefont {T.}~\bibnamefont {Liang}}, \bibinfo {author} {\bibfnamefont {K.}~\bibnamefont {Xu}}, \bibinfo {author} {\bibfnamefont {E.}~\bibnamefont {Lindgren}}, \bibinfo {author} {\bibfnamefont {Z.}~\bibnamefont {Chen}}, \bibinfo {author} {\bibfnamefont {R.}~\bibnamefont {Zhao}}, \bibinfo {author} {\bibfnamefont {J.}~\bibnamefont {Liu}}, \bibinfo {author} {\bibfnamefont {B.}~\bibnamefont {Tang}}, \bibinfo {author} {\bibfnamefont {B.}~\bibnamefont {Zhang}}, \bibinfo {author} {\bibfnamefont {Y.}~\bibnamefont {Wang}}, \bibinfo {author} {\bibfnamefont {K.}~\bibnamefont {Song}}, \bibinfo {author} {\bibfnamefont {P.}~\bibnamefont {Ying}}, \bibinfo {author} {\bibfnamefont {H.}~\bibnamefont {Dong}}, \bibinfo {author} {\bibfnamefont {S.}~\bibnamefont {Chen}}, \bibinfo {author} {\bibfnamefont {P.}~\bibnamefont {Erhart}}, \bibinfo {author} {\bibfnamefont {Z.}~\bibnamefont {Fan}}, \bibinfo {author} {\bibfnamefont {T.}~\bibnamefont {Ala-Nissila}},\ and\ \bibinfo {author} {\bibfnamefont {J.}~\bibnamefont {Xu}},\ }\href {https://arxiv.org/abs/2504.21286} {\bibinfo {title} {Nep89: Universal neuroevolution potential for inorganic and organic materials across 89 elements}} (\bibinfo {year} {2025}),\ \Eprint {https://arxiv.org/abs/2504.21286} {arXiv:2504.21286 [cond-mat.mtrl-sci]} \BibitemShut {NoStop}%
\bibitem [{\citenamefont {Fan}(2025)}]{nepcpu}%
  \BibitemOpen
  \bibfield  {author} {\bibinfo {author} {\bibfnamefont {Z.}~\bibnamefont {Fan}},\ }\href@noop {} {\bibinfo {title} {{NEP\_CPU}}},\ \bibinfo {howpublished} {\url{https://github.com/brucefan1983/NEP_CPU}} (\bibinfo {year} {2025})\BibitemShut {NoStop}%
\bibitem [{\citenamefont {Hjorth~Larsen}\ \emph {et~al.}(2017)\citenamefont {Hjorth~Larsen}, \citenamefont {J{\o}rgen~Mortensen}, \citenamefont {Blomqvist}, \citenamefont {Castelli}, \citenamefont {Christensen}, \citenamefont {Du{\l}ak}, \citenamefont {Friis}, \citenamefont {Groves}, \citenamefont {Hammer}, \citenamefont {Hargus}, \citenamefont {Hermes}, \citenamefont {Jennings}, \citenamefont {Bjerre~Jensen}, \citenamefont {Kermode}, \citenamefont {Kitchin}, \citenamefont {Leonhard~Kolsbjerg}, \citenamefont {Kubal}, \citenamefont {Kaasbjerg}, \citenamefont {Lysgaard}, \citenamefont {Bergmann~Maronsson}, \citenamefont {Maxson}, \citenamefont {Olsen}, \citenamefont {Pastewka}, \citenamefont {Peterson}, \citenamefont {Rostgaard}, \citenamefont {Schi{\o}tz}, \citenamefont {Sch{\"u}tt}, \citenamefont {Strange}, \citenamefont {Thygesen}, \citenamefont {Vegge}, \citenamefont {Vilhelmsen}, \citenamefont {Walter}, \citenamefont {Zeng},\ and\ \citenamefont {Jacobsen}}]{Hjorthlarsen2017TheAtomic}%
  \BibitemOpen
  \bibfield  {author} {\bibinfo {author} {\bibfnamefont {A.}~\bibnamefont {Hjorth~Larsen}}, \bibinfo {author} {\bibfnamefont {J.}~\bibnamefont {J{\o}rgen~Mortensen}}, \bibinfo {author} {\bibfnamefont {J.}~\bibnamefont {Blomqvist}}, \bibinfo {author} {\bibfnamefont {I.~E.}\ \bibnamefont {Castelli}}, \bibinfo {author} {\bibfnamefont {R.}~\bibnamefont {Christensen}}, \bibinfo {author} {\bibfnamefont {M.}~\bibnamefont {Du{\l}ak}}, \bibinfo {author} {\bibfnamefont {J.}~\bibnamefont {Friis}}, \bibinfo {author} {\bibfnamefont {M.~N.}\ \bibnamefont {Groves}}, \bibinfo {author} {\bibfnamefont {B.}~\bibnamefont {Hammer}}, \bibinfo {author} {\bibfnamefont {C.}~\bibnamefont {Hargus}}, \bibinfo {author} {\bibfnamefont {E.~D.}\ \bibnamefont {Hermes}}, \bibinfo {author} {\bibfnamefont {P.~C.}\ \bibnamefont {Jennings}}, \bibinfo {author} {\bibfnamefont {P.}~\bibnamefont {Bjerre~Jensen}}, \bibinfo {author} {\bibfnamefont {J.}~\bibnamefont {Kermode}}, \bibinfo {author} {\bibfnamefont {J.~R.}\ \bibnamefont {Kitchin}}, \bibinfo {author} {\bibfnamefont {E.}~\bibnamefont {Leonhard~Kolsbjerg}}, \bibinfo {author} {\bibfnamefont {J.}~\bibnamefont {Kubal}}, \bibinfo {author} {\bibfnamefont {K.}~\bibnamefont {Kaasbjerg}}, \bibinfo {author} {\bibfnamefont {S.}~\bibnamefont {Lysgaard}}, \bibinfo {author} {\bibfnamefont {J.}~\bibnamefont {Bergmann~Maronsson}}, \bibinfo {author} {\bibfnamefont {T.}~\bibnamefont {Maxson}}, \bibinfo {author} {\bibfnamefont {T.}~\bibnamefont {Olsen}}, \bibinfo {author} {\bibfnamefont {L.}~\bibnamefont {Pastewka}}, \bibinfo {author} {\bibfnamefont {A.}~\bibnamefont {Peterson}}, \bibinfo {author} {\bibfnamefont {C.}~\bibnamefont {Rostgaard}}, \bibinfo {author} {\bibfnamefont {J.}~\bibnamefont {Schi{\o}tz}}, \bibinfo {author} {\bibfnamefont {O.}~\bibnamefont {Sch{\"u}tt}}, \bibinfo {author} {\bibfnamefont {M.}~\bibnamefont {Strange}}, \bibinfo {author} {\bibfnamefont {K.~S.}\ \bibnamefont {Thygesen}}, \bibinfo {author} {\bibfnamefont {T.}~\bibnamefont {Vegge}}, \bibinfo {author} {\bibfnamefont {L.}~\bibnamefont {Vilhelmsen}}, \bibinfo {author} {\bibfnamefont {M.}~\bibnamefont {Walter}}, \bibinfo {author} {\bibfnamefont {Z.}~\bibnamefont {Zeng}},\ and\ \bibinfo {author} {\bibfnamefont {K.~W.}\ \bibnamefont {Jacobsen}},\ }\bibfield  {title} {\bibinfo {title} {The atomic simulation environment---a {{Python}} library for working with atoms},\ }\href {https://doi.org/10.1088/1361-648X/aa680e} {\bibfield  {journal} {\bibinfo  {journal} {Journal of Physics: Condensed Matter}\ }\textbf {\bibinfo {volume} {29}},\ \bibinfo {pages} {273002} (\bibinfo {year} {2017})}\BibitemShut {NoStop}%
\bibitem [{\citenamefont {J{\"a}ger}\ \emph {et~al.}(2018)\citenamefont {J{\"a}ger}, \citenamefont {Morooka}, \citenamefont {Federici~Canova}, \citenamefont {Himanen},\ and\ \citenamefont {Foster}}]{Jager2018Machine}%
  \BibitemOpen
  \bibfield  {author} {\bibinfo {author} {\bibfnamefont {M.~O.~J.}\ \bibnamefont {J{\"a}ger}}, \bibinfo {author} {\bibfnamefont {E.~V.}\ \bibnamefont {Morooka}}, \bibinfo {author} {\bibfnamefont {F.}~\bibnamefont {Federici~Canova}}, \bibinfo {author} {\bibfnamefont {L.}~\bibnamefont {Himanen}},\ and\ \bibinfo {author} {\bibfnamefont {A.~S.}\ \bibnamefont {Foster}},\ }\bibfield  {title} {\bibinfo {title} {Machine learning hydrogen adsorption on nanoclusters through structural descriptors},\ }\href {https://doi.org/10.1038/s41524-018-0096-5} {\bibfield  {journal} {\bibinfo  {journal} {npj Computational Materials}\ }\textbf {\bibinfo {volume} {4}},\ \bibinfo {pages} {37} (\bibinfo {year} {2018})}\BibitemShut {NoStop}%
\bibitem [{\citenamefont {Bart{\'o}k}\ \emph {et~al.}(2017)\citenamefont {Bart{\'o}k}, \citenamefont {De}, \citenamefont {Poelking}, \citenamefont {Bernstein}, \citenamefont {Kermode}, \citenamefont {Cs{\'a}nyi},\ and\ \citenamefont {Ceriotti}}]{Bartok2017Machine}%
  \BibitemOpen
  \bibfield  {author} {\bibinfo {author} {\bibfnamefont {A.~P.}\ \bibnamefont {Bart{\'o}k}}, \bibinfo {author} {\bibfnamefont {S.}~\bibnamefont {De}}, \bibinfo {author} {\bibfnamefont {C.}~\bibnamefont {Poelking}}, \bibinfo {author} {\bibfnamefont {N.}~\bibnamefont {Bernstein}}, \bibinfo {author} {\bibfnamefont {J.~R.}\ \bibnamefont {Kermode}}, \bibinfo {author} {\bibfnamefont {G.}~\bibnamefont {Cs{\'a}nyi}},\ and\ \bibinfo {author} {\bibfnamefont {M.}~\bibnamefont {Ceriotti}},\ }\bibfield  {title} {\bibinfo {title} {Machine learning unifies the modeling of materials and molecules},\ }\href {https://doi.org/10.1126/sciadv.1701816} {\bibfield  {journal} {\bibinfo  {journal} {Science Advances}\ }\textbf {\bibinfo {volume} {3}},\ \bibinfo {pages} {e1701816} (\bibinfo {year} {2017})}\BibitemShut {NoStop}%
\bibitem [{\citenamefont {Himanen}\ \emph {et~al.}(2020)\citenamefont {Himanen}, \citenamefont {J{\"a}ger}, \citenamefont {Morooka}, \citenamefont {Federici~Canova}, \citenamefont {Ranawat}, \citenamefont {Gao}, \citenamefont {Rinke},\ and\ \citenamefont {Foster}}]{Himanen2020DScribe}%
  \BibitemOpen
  \bibfield  {author} {\bibinfo {author} {\bibfnamefont {L.}~\bibnamefont {Himanen}}, \bibinfo {author} {\bibfnamefont {M.~O.}\ \bibnamefont {J{\"a}ger}}, \bibinfo {author} {\bibfnamefont {E.~V.}\ \bibnamefont {Morooka}}, \bibinfo {author} {\bibfnamefont {F.}~\bibnamefont {Federici~Canova}}, \bibinfo {author} {\bibfnamefont {Y.~S.}\ \bibnamefont {Ranawat}}, \bibinfo {author} {\bibfnamefont {D.~Z.}\ \bibnamefont {Gao}}, \bibinfo {author} {\bibfnamefont {P.}~\bibnamefont {Rinke}},\ and\ \bibinfo {author} {\bibfnamefont {A.~S.}\ \bibnamefont {Foster}},\ }\bibfield  {title} {\bibinfo {title} {{{DScribe}}: {{Library}} of descriptors for machine learning in materials science},\ }\href {https://doi.org/10.1016/j.cpc.2019.106949} {\bibfield  {journal} {\bibinfo  {journal} {Computer Physics Communications}\ }\textbf {\bibinfo {volume} {247}},\ \bibinfo {pages} {106949} (\bibinfo {year} {2020})}\BibitemShut {NoStop}%
\bibitem [{\citenamefont {Pedregosa}\ \emph {et~al.}(2011)\citenamefont {Pedregosa}, \citenamefont {Varoquaux}, \citenamefont {Gramfort}, \citenamefont {Michel}, \citenamefont {Thirion}, \citenamefont {Grisel}, \citenamefont {Blondel}, \citenamefont {Prettenhofer}, \citenamefont {Weiss}, \citenamefont {Dubourg}, \citenamefont {Vanderplas}, \citenamefont {Passos}, \citenamefont {Cournapeau}, \citenamefont {Brucher}, \citenamefont {Perrot},\ and\ \citenamefont {Duchesnay}}]{Pedregosa2011Scikitlearn}%
  \BibitemOpen
  \bibfield  {author} {\bibinfo {author} {\bibfnamefont {F.}~\bibnamefont {Pedregosa}}, \bibinfo {author} {\bibfnamefont {G.}~\bibnamefont {Varoquaux}}, \bibinfo {author} {\bibfnamefont {A.}~\bibnamefont {Gramfort}}, \bibinfo {author} {\bibfnamefont {V.}~\bibnamefont {Michel}}, \bibinfo {author} {\bibfnamefont {B.}~\bibnamefont {Thirion}}, \bibinfo {author} {\bibfnamefont {O.}~\bibnamefont {Grisel}}, \bibinfo {author} {\bibfnamefont {M.}~\bibnamefont {Blondel}}, \bibinfo {author} {\bibfnamefont {P.}~\bibnamefont {Prettenhofer}}, \bibinfo {author} {\bibfnamefont {R.}~\bibnamefont {Weiss}}, \bibinfo {author} {\bibfnamefont {V.}~\bibnamefont {Dubourg}}, \bibinfo {author} {\bibfnamefont {J.}~\bibnamefont {Vanderplas}}, \bibinfo {author} {\bibfnamefont {A.}~\bibnamefont {Passos}}, \bibinfo {author} {\bibfnamefont {D.}~\bibnamefont {Cournapeau}}, \bibinfo {author} {\bibfnamefont {M.}~\bibnamefont {Brucher}}, \bibinfo {author} {\bibfnamefont {M.}~\bibnamefont {Perrot}},\ and\ \bibinfo {author} {\bibfnamefont {{\'E}.}~\bibnamefont {Duchesnay}},\ }\bibfield  {title} {\bibinfo {title} {Scikit-learn: {{Machine Learning}} in {{Python}}},\ }\href@noop {} {\bibfield  {journal} {\bibinfo  {journal} {Journal of Machine Learning Research}\ }\textbf {\bibinfo {volume} {12}},\ \bibinfo {pages} {2825} (\bibinfo {year} {2011})}\BibitemShut {NoStop}%
\bibitem [{\citenamefont {McInnes}\ \emph {et~al.}(2020)\citenamefont {McInnes}, \citenamefont {Healy},\ and\ \citenamefont {Melville}}]{Mcinnes2020UMAP}%
  \BibitemOpen
  \bibfield  {author} {\bibinfo {author} {\bibfnamefont {L.}~\bibnamefont {McInnes}}, \bibinfo {author} {\bibfnamefont {J.}~\bibnamefont {Healy}},\ and\ \bibinfo {author} {\bibfnamefont {J.}~\bibnamefont {Melville}},\ }\href {https://doi.org/10.48550/arXiv.1802.03426} {\bibinfo {title} {{{UMAP}}: {{Uniform}} manifold approximation and projection for dimension reduction}} (\bibinfo {year} {2020}),\ \Eprint {https://arxiv.org/abs/1802.03426} {arXiv:1802.03426 [stat]} \BibitemShut {NoStop}%
\bibitem [{\citenamefont {Stukowski}(2010)}]{Stukowski2010Visualization}%
  \BibitemOpen
  \bibfield  {author} {\bibinfo {author} {\bibfnamefont {A.}~\bibnamefont {Stukowski}},\ }\bibfield  {title} {\bibinfo {title} {Visualization and analysis of atomistic simulation data with {{OVITO}}--the open visualization tool},\ }\href {https://doi.org/10.1088/0965-0393/18/1/015012} {\bibfield  {journal} {\bibinfo  {journal} {Modelling and Simulation in Materials Science and Engineering}\ }\textbf {\bibinfo {volume} {18}},\ \bibinfo {pages} {015012} (\bibinfo {year} {2010})}\BibitemShut {NoStop}%
\bibitem [{\citenamefont {Humphrey}\ \emph {et~al.}(1996)\citenamefont {Humphrey}, \citenamefont {Dalke},\ and\ \citenamefont {Schulten}}]{William1996VMD}%
  \BibitemOpen
  \bibfield  {author} {\bibinfo {author} {\bibfnamefont {W.}~\bibnamefont {Humphrey}}, \bibinfo {author} {\bibfnamefont {A.}~\bibnamefont {Dalke}},\ and\ \bibinfo {author} {\bibfnamefont {K.}~\bibnamefont {Schulten}},\ }\bibfield  {title} {\bibinfo {title} {Vmd: Visual molecular dynamics},\ }\href {https://doi.org/https://doi.org/10.1016/0263-7855(96)00018-5} {\bibfield  {journal} {\bibinfo  {journal} {Journal of Molecular Graphics}\ }\textbf {\bibinfo {volume} {14}},\ \bibinfo {pages} {33} (\bibinfo {year} {1996})}\BibitemShut {NoStop}%
\bibitem [{\citenamefont {Fraux}\ \emph {et~al.}(2020)\citenamefont {Fraux}, \citenamefont {Cersonsky},\ and\ \citenamefont {Ceriotti}}]{fraux2020Chemiscope}%
  \BibitemOpen
  \bibfield  {author} {\bibinfo {author} {\bibfnamefont {G.}~\bibnamefont {Fraux}}, \bibinfo {author} {\bibfnamefont {R.~K.}\ \bibnamefont {Cersonsky}},\ and\ \bibinfo {author} {\bibfnamefont {M.}~\bibnamefont {Ceriotti}},\ }\bibfield  {title} {\bibinfo {title} {Chemiscope: {{Interactive}} structure-property explorer for materials and molecules},\ }\href {https://doi.org/10.21105/joss.02117} {\bibfield  {journal} {\bibinfo  {journal} {Journal of Open Source Software}\ }\textbf {\bibinfo {volume} {5}},\ \bibinfo {pages} {2117} (\bibinfo {year} {2020})}\BibitemShut {NoStop}%
\bibitem [{\citenamefont {Campagnola}\ \emph {et~al.}(2025)\citenamefont {Campagnola}, \citenamefont {Larson}, \citenamefont {Klein}, \citenamefont {Hoese}, \citenamefont {{Siddharth}}, \citenamefont {Rossant}, \citenamefont {Griffiths}, \citenamefont {Rougier}, \citenamefont {{asnt}}, \citenamefont {Gaifas}, \citenamefont {M{\"u}hlbauer}, \citenamefont {Taylor}, \citenamefont {{MSS}}, \citenamefont {Lambert}, \citenamefont {{sylm21}}, \citenamefont {Anderson}, \citenamefont {Champandard}, \citenamefont {Hunter}, \citenamefont {Robitaille}, \citenamefont {Kaptan}, \citenamefont {{de Andrade}}, \citenamefont {Bokota}, \citenamefont {Favelier}, \citenamefont {Harfouche}, \citenamefont {Combrisson}, \citenamefont {{ThenTech}}, \citenamefont {{fschill}}, \citenamefont {Bacchini},\ and\ \citenamefont {Aye}}]{Campagnola2025Vispy}%
  \BibitemOpen
  \bibfield  {author} {\bibinfo {author} {\bibfnamefont {L.}~\bibnamefont {Campagnola}}, \bibinfo {author} {\bibfnamefont {E.}~\bibnamefont {Larson}}, \bibinfo {author} {\bibfnamefont {A.}~\bibnamefont {Klein}}, \bibinfo {author} {\bibfnamefont {D.}~\bibnamefont {Hoese}}, \bibinfo {author} {\bibnamefont {{Siddharth}}}, \bibinfo {author} {\bibfnamefont {C.}~\bibnamefont {Rossant}}, \bibinfo {author} {\bibfnamefont {A.}~\bibnamefont {Griffiths}}, \bibinfo {author} {\bibfnamefont {N.~P.}\ \bibnamefont {Rougier}}, \bibinfo {author} {\bibnamefont {{asnt}}}, \bibinfo {author} {\bibfnamefont {L.}~\bibnamefont {Gaifas}}, \bibinfo {author} {\bibfnamefont {K.}~\bibnamefont {M{\"u}hlbauer}}, \bibinfo {author} {\bibfnamefont {A.}~\bibnamefont {Taylor}}, \bibinfo {author} {\bibnamefont {{MSS}}}, \bibinfo {author} {\bibfnamefont {T.}~\bibnamefont {Lambert}}, \bibinfo {author} {\bibnamefont {{sylm21}}}, \bibinfo {author} {\bibfnamefont {A.}~\bibnamefont {Anderson}}, \bibinfo {author} {\bibfnamefont {A.~J.}\ \bibnamefont {Champandard}}, \bibinfo {author} {\bibfnamefont {M.}~\bibnamefont {Hunter}}, \bibinfo {author} {\bibfnamefont {T.}~\bibnamefont {Robitaille}}, \bibinfo {author} {\bibfnamefont {M.~F.}\ \bibnamefont {Kaptan}}, \bibinfo {author} {\bibfnamefont {E.~S.}\ \bibnamefont {{de Andrade}}}, \bibinfo {author} {\bibfnamefont {G.}~\bibnamefont {Bokota}}, \bibinfo {author} {\bibfnamefont {G.}~\bibnamefont {Favelier}}, \bibinfo {author} {\bibfnamefont {M.}~\bibnamefont {Harfouche}}, \bibinfo {author} {\bibfnamefont {E.}~\bibnamefont {Combrisson}}, \bibinfo {author} {\bibnamefont {{ThenTech}}}, \bibinfo {author} {\bibnamefont {{fschill}}}, \bibinfo {author} {\bibfnamefont {A.}~\bibnamefont {Bacchini}},\ and\ \bibinfo {author} {\bibfnamefont {M.}~\bibnamefont {Aye}},\ }\href {https://doi.org/10.5281/zenodo.15263124} {\bibinfo {title} {Vispy/vispy: {{V0}}.15.0}},\ \bibinfo {howpublished} {Zenodo} (\bibinfo {year} {2025})\BibitemShut {NoStop}%
\bibitem [{\citenamefont {Xu}\ \emph {et~al.}(2025)\citenamefont {Xu}, \citenamefont {Wu}, \citenamefont {Bai}, \citenamefont {Wang}, \citenamefont {Zhou}, \citenamefont {Fan}, \citenamefont {Zhang}, \citenamefont {Tan}, \citenamefont {Li}, \citenamefont {Bian},\ and\ \citenamefont {Liu}}]{Xu2025Recordefficiency}%
  \BibitemOpen
  \bibfield  {author} {\bibinfo {author} {\bibfnamefont {D.}~\bibnamefont {Xu}}, \bibinfo {author} {\bibfnamefont {M.}~\bibnamefont {Wu}}, \bibinfo {author} {\bibfnamefont {Y.}~\bibnamefont {Bai}}, \bibinfo {author} {\bibfnamefont {B.}~\bibnamefont {Wang}}, \bibinfo {author} {\bibfnamefont {H.}~\bibnamefont {Zhou}}, \bibinfo {author} {\bibfnamefont {Z.}~\bibnamefont {Fan}}, \bibinfo {author} {\bibfnamefont {N.}~\bibnamefont {Zhang}}, \bibinfo {author} {\bibfnamefont {J.}~\bibnamefont {Tan}}, \bibinfo {author} {\bibfnamefont {H.}~\bibnamefont {Li}}, \bibinfo {author} {\bibfnamefont {H.}~\bibnamefont {Bian}},\ and\ \bibinfo {author} {\bibfnamefont {Z.}~\bibnamefont {Liu}},\ }\bibfield  {title} {\bibinfo {title} {Record-efficiency inverted {{CsPbI3}} perovskite solar cells enabled by rearrangement and hydrophilic modification of {{SAMs}}},\ }\href {https://doi.org/10.1002/adfm.202412946} {\bibfield  {journal} {\bibinfo  {journal} {Advanced Functional Materials}\ }\textbf {\bibinfo {volume} {35}},\ \bibinfo {pages} {2412946} (\bibinfo {year} {2025})}\BibitemShut {NoStop}%
\bibitem [{\citenamefont {Marronnier}\ \emph {et~al.}(2018)\citenamefont {Marronnier}, \citenamefont {Roma}, \citenamefont {{Boyer-Richard}}, \citenamefont {Pedesseau}, \citenamefont {Jancu}, \citenamefont {Bonnassieux}, \citenamefont {Katan}, \citenamefont {Stoumpos}, \citenamefont {Kanatzidis},\ and\ \citenamefont {Even}}]{Marronnier2018Anharmonicity}%
  \BibitemOpen
  \bibfield  {author} {\bibinfo {author} {\bibfnamefont {A.}~\bibnamefont {Marronnier}}, \bibinfo {author} {\bibfnamefont {G.}~\bibnamefont {Roma}}, \bibinfo {author} {\bibfnamefont {S.}~\bibnamefont {{Boyer-Richard}}}, \bibinfo {author} {\bibfnamefont {L.}~\bibnamefont {Pedesseau}}, \bibinfo {author} {\bibfnamefont {J.-M.}\ \bibnamefont {Jancu}}, \bibinfo {author} {\bibfnamefont {Y.}~\bibnamefont {Bonnassieux}}, \bibinfo {author} {\bibfnamefont {C.}~\bibnamefont {Katan}}, \bibinfo {author} {\bibfnamefont {C.~C.}\ \bibnamefont {Stoumpos}}, \bibinfo {author} {\bibfnamefont {M.~G.}\ \bibnamefont {Kanatzidis}},\ and\ \bibinfo {author} {\bibfnamefont {J.}~\bibnamefont {Even}},\ }\bibfield  {title} {\bibinfo {title} {Anharmonicity and disorder in the black phases of cesium lead iodide used for stable inorganic perovskite solar cells},\ }\href {https://doi.org/10.1021/acsnano.8b00267} {\bibfield  {journal} {\bibinfo  {journal} {ACS Nano}\ }\textbf {\bibinfo {volume} {12}},\ \bibinfo {pages} {3477} (\bibinfo {year} {2018})}\BibitemShut {NoStop}%
\bibitem [{\citenamefont {Momma}\ and\ \citenamefont {Izumi}(2011)}]{momma2011VESTA3}%
  \BibitemOpen
  \bibfield  {author} {\bibinfo {author} {\bibfnamefont {K.}~\bibnamefont {Momma}}\ and\ \bibinfo {author} {\bibfnamefont {F.}~\bibnamefont {Izumi}},\ }\bibfield  {title} {\bibinfo {title} {{{{\emph{VESTA}}}}{\emph{ 3}} for three-dimensional visualization of crystal, volumetric and morphology data},\ }\href {https://doi.org/10.1107/S0021889811038970} {\bibfield  {journal} {\bibinfo  {journal} {Journal of Applied Crystallography}\ }\textbf {\bibinfo {volume} {44}},\ \bibinfo {pages} {1272} (\bibinfo {year} {2011})}\BibitemShut {NoStop}%
\bibitem [{\citenamefont {Bl{\"o}chl}(1994)}]{Blochl1994Projector}%
  \BibitemOpen
  \bibfield  {author} {\bibinfo {author} {\bibfnamefont {P.~E.}\ \bibnamefont {Bl{\"o}chl}},\ }\bibfield  {title} {\bibinfo {title} {Projector augmented-wave method},\ }\href {https://doi.org/10.1103/PhysRevB.50.17953} {\bibfield  {journal} {\bibinfo  {journal} {Physical Review B}\ }\textbf {\bibinfo {volume} {50}},\ \bibinfo {pages} {17953} (\bibinfo {year} {1994})}\BibitemShut {NoStop}%
\bibitem [{\citenamefont {Perdew}\ \emph {et~al.}(1996)\citenamefont {Perdew}, \citenamefont {Burke},\ and\ \citenamefont {Ernzerhof}}]{Perdew1996Generalized}%
  \BibitemOpen
  \bibfield  {author} {\bibinfo {author} {\bibfnamefont {J.~P.}\ \bibnamefont {Perdew}}, \bibinfo {author} {\bibfnamefont {K.}~\bibnamefont {Burke}},\ and\ \bibinfo {author} {\bibfnamefont {M.}~\bibnamefont {Ernzerhof}},\ }\bibfield  {title} {\bibinfo {title} {Generalized {{Gradient Approximation Made Simple}}},\ }\href {https://doi.org/10.1103/PhysRevLett.77.3865} {\bibfield  {journal} {\bibinfo  {journal} {Physical Review Letters}\ }\textbf {\bibinfo {volume} {77}},\ \bibinfo {pages} {3865} (\bibinfo {year} {1996})}\BibitemShut {NoStop}%
\bibitem [{\citenamefont {Bernetti}\ and\ \citenamefont {Bussi}(2020)}]{Bernetti2020Pressure}%
  \BibitemOpen
  \bibfield  {author} {\bibinfo {author} {\bibfnamefont {M.}~\bibnamefont {Bernetti}}\ and\ \bibinfo {author} {\bibfnamefont {G.}~\bibnamefont {Bussi}},\ }\bibfield  {title} {\bibinfo {title} {Pressure control using stochastic cell rescaling},\ }\href {https://doi.org/10.1063/5.0020514} {\bibfield  {journal} {\bibinfo  {journal} {Journal of Chemical Physics}\ }\textbf {\bibinfo {volume} {153}},\ \bibinfo {pages} {114107} (\bibinfo {year} {2020})}\BibitemShut {NoStop}%
\end{thebibliography}%

\end{document}